\documentclass[10pt]{article} 
\usepackage[accepted]{rlc}

\usepackage{amssymb}            
\usepackage{mathtools}          
\usepackage{mathrsfs}           
\usepackage{graphicx}           
\usepackage{subcaption}         
\usepackage[space]{grffile}     
\usepackage{url}                

\usepackage{amsmath,amsfonts}
\usepackage{algorithmic}
\usepackage{algorithm}
\usepackage{array}
\usepackage{textcomp}
\usepackage{stfloats}
\usepackage{verbatim}
\usepackage{times}
\usepackage{soul}

\usepackage{booktabs}
\usepackage{xcolor}
\usepackage[switch]{lineno}
\usepackage{bbm}

\usepackage{tikz}
\usetikzlibrary{positioning}
\usepackage{multirow}

\usepackage{caption}
\usepackage{subcaption}
\usepackage{epstopdf}

\newcommand{\keypoint}[1]{\noindent \textbf{#1}\quad}

\title{Enabling Intelligent Interactions between an Agent and an LLM: A Reinforcement Learning Approach}

\author{Bin Hu\thanks{Equal contributions}\\
    hubin@@zhejianglab.com \\
    Zhejiang Lab
    \And
    Chenyang Zhao$^{*}$ \\
    c.zhao@zhejianglab.com\\
    Zhejiang Lab
    \And
    Pu Zhang$^{*}$ \\
    puz@zhejianglab.com\\
    Zhejiang Lab
    \And
    Zihao Zhou$^{*}$ \\
    zhouzihao@zhejianglab.com\\
    Zhejiang Lab
    \And
    Yuanhang Yang\thanks{Y. Yang did this work during internship at Zhejiang Lab.} \\
    ysngkil@gmail.com\\
    Harbin Institute of Technology (Shenzhen)
    \And
    Zenglin Xu \\
    zenglin@gmail.com\\
    AI Innovation and Incubation Institute, Fudan University
    \And
    Bin Liu\thanks{Corresponding author} \\
    bins@ieee.org\\
    Zhejiang Lab
}

\begin{document}

\maketitle

\begin{abstract}
Large language models (LLMs) encode a vast amount of world knowledge acquired from massive text datasets. Recent studies have demonstrated that LLMs can assist an embodied agent in solving complex sequential decision making tasks by providing high-level instructions. However, interactions with LLMs can be time-consuming. In many practical scenarios, it requires a significant amount of storage space that can only be deployed on remote cloud servers. Additionally, using commercial LLMs can be costly since they may charge based on usage frequency. In this paper, we explore how to enable intelligent cost-effective interactions between a down stream task oriented agent and an LLM. We find that this problem can be naturally formulated by a Markov decision process (MDP), and propose \textit{When2Ask}, a reinforcement learning based approach that learns when it is necessary to query LLMs for high-level instructions to accomplish a target task. On one side, \textit{When2Ask} discourages unnecessary redundant interactions, while on the other side, it enables the agent to identify and follow useful instructions from the LLM. This enables the agent to halt an ongoing plan and transition to a more suitable one based on new environmental observations. Experiments on MiniGrid and Habitat environments that entail planning sub-goals demonstrate that \textit{When2Ask} learns to solve target tasks with only a few necessary interactions with the LLM, significantly reducing interaction costs in testing environments compared with baseline methods. Our code is available at: https://github.com/ZJLAB-AMMI/LLM4RL.
\end{abstract}

\section{Introduction}
\label{sec: intro}
\begin{figure}[t]
    \centering
    \includegraphics[width=0.49 \textwidth]{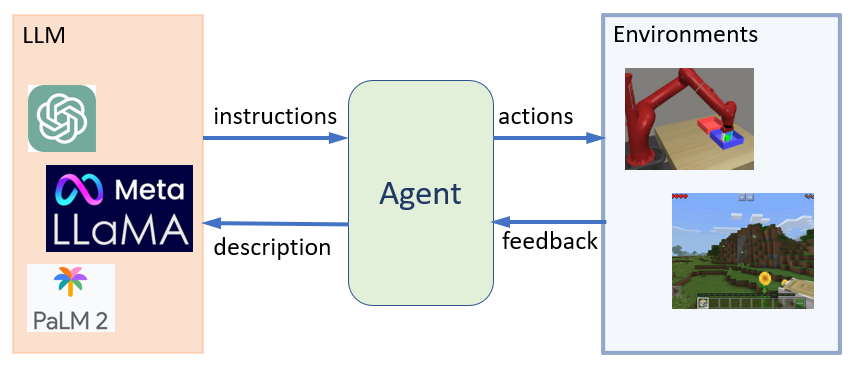}
    \caption{A general framework of using LLMs for solving complex embodied tasks. The LLMs provide high-level instructions based on state descriptions, and the agent generates low-level actions following these instructions and interacts with the target environment to collect further feedback.}
    \label{fig: overview}
\end{figure}

To empower embodied agents with the capability to effectively handle demanding sequential decision-making tasks, it is essential for them to possess reasoning abilities that enable them to plan for the long-term consequences of their actions \cite{deitke2022retrospectives}. Reinforcement learning (RL), particularly deep RL, has emerged as a popular paradigm for addressing these challenges. Deep RL involves agents interacting with the environment and learning from feedback to improve their decision-making over time.
Despite recent advancements, several challenges still remains and limits its vast applications in real world scenarios.
For instance, solving complex problems using deep RL often requires significant computational resources. Additionally, safety concerns can arise during the learning phase, especially in scenarios where the agent's exploration might interact with the real world or other sensitive environments \cite{das2018embodied,chevalier2018babyai}.
As an alternative, the emergence of large language models (LLMs) has shown promise in tackling these issues. Previous studies have demonstrated that LLMs possess reasoning capabilities \cite{radford2019language,brown2020language,wei2022chain}. Researchers have explored leveraging LLMs' reasoning abilities to solve various embodied tasks, including robot manipulation tasks \cite{ahn2022can,huang2022inner,jiang2022vima} and playing video games \cite{dasgupta2023collaborating,wang2023voyager,wang2023describe}. As depicted in Fig.~\ref{fig: overview}, the embodied agent interacts with the environment, gathering information ralated to the target task, and utilizes LLMs as explicit reasoners to make high-level plans using natural language instructions, such as instructing a robot to ``pick up a can of coke'' or ``place an apple on the table''.

While the integration of pre-trained LLMs as explicit planners in embodied agents has demonstrated promising results, enabling efficient interaction between these agents and LLMs to solve real-world problems remains challenging. Frequent queries to LLMs can result in unnecessary resource wastage, including fees (if a commercial LLM is used), communication overhead and reasoning time. Whereas insufficient queries to LLMs prevent the agent from obtaining useful instructions in time to adjust its plan to respond to the complex and changing environment.

Determining an appropriate guideline for querying LLMs requires expert knowledge of the target task.
Consider a scenario where a robot is instructed to collect a can of coke but encounters a locked door on its way to the kitchen. Ideally, the agent should recognize this incident and adjust its plan accordingly by consulting the LLM on how to deal with the locked door.
In such cases, timely decision-making regarding when to consult the LLM planner becomes crucial. Failure to interrupt the ongoing action plan and request a new one in time can hinder task completion progress or even lead to safety issues, such as damaging the door or the robot itself. Conversely, frequent requests for plans from the LLM can be time-consuming and costly, particularly when using commercial LLMs deployed on remote cloud servers that charge based on usage frequency.

In this paper, we propose \textit{When2Ask}, a general approach that trains the agent to make intelligent cost-effective interactions between itself and an LLM remotely deployed. Our objective is to facilitate effective completion of a target task while reducing unnecessary non-informative interactions with the LLM. Specifically, we adopt a Planner-Actor-Mediator framework, similar to \cite{dasgupta2023collaborating}, where the planner is a pre-trained LLM used for making plans, the actor contains policies for executing the plans, and the mediator serves as an interface in between by deciding when to request a new plan and generating observation representations for the planner (which are text descriptions). With a focus on optimizing interacting timings, we use RL to learn an asking policy that instructs the agent to either adhere to the current plan or request a new plan from the LLM.

To summarize, our \emph{\textbf{main contributions}} include:
\begin{itemize}
    \item We propose an RL approach termed \textit{When2Ask} to coordinate the interaction between a down stream task oriented agent and a pre-trained LLM based on the Planner-Actor-Mediator framework \cite{dasgupta2023collaborating}. Specifically, we propose to introduce an explicit asking policy in the mediator and train it using an RL approach to determine when to query the LLM planner.
    \item We conducted a comprehensive evaluation of \textit{When2Ask} against baseline methods based on simulation platforms MiniGrid \cite{minigrid} and Habitat \cite{szot2021habitat}. The results demonstrate that the learned asking policy is able to make intelligent decisions on when to query LLMs, resulting in high success rates with only a few necessary LLM interactions in the testing phase. In contrast to conventional interaction strategies that rely on predetermined termination criteria, \textit{When2Ask} offers a significant advantage by enabling the interruption of an ongoing plan in favor of a new plan that addresses emerging observations.
\end{itemize}

\section{Preliminary}
\label{sec: preliminary}
\subsection{The Options Framework}
We consider sequential decision-making in embodied environments, which is commonly formalized as a Markov decision process (MDP), denoted as $\mathcal{M} = \langle \mathcal{S}, \mathcal{A}, p, r, \gamma \rangle$. Here $\mathcal{S}$ represents the state space, $\mathcal{A}$ represents the action spaces, $p(s'|s,a)$ denotes the state transition probability function, $r(s,a)$ represents the reward function, and $\gamma$ is the discount factor. The objective is to learn an optimal policy that maximizes the cumulative reward over time $\sum_t \gamma^t r(s_t,a_t)$, where $t$ denotes the  time index.

The options framework extends the traditional notion of action in an MDP to include options, which are essentially closed-loop policies that encompass a sequence of actions over a period of time \cite{sutton1999between, precup2000temporal}. Options can range from higher-level tasks such as picking up an object or going to lunch, to more primitive actions like muscle twitches and joint torques. The introduction of options allows for the incorporation of temporally abstract knowledge and action within the RL framework in a natural and general manner, thus provides a flexible and intuitive approach to handle complex tasks with varying levels of granularity. Formally, an option $\omega$ is defined as a 3-tuple $\langle \mathcal{I_\omega}, \pi_\omega, \beta_\omega \rangle,$ where $\mathcal{I}_\omega$ represents the initial state set, $\pi_\omega$ denotes the acting policy, and $\beta_\omega$ represents the termination condition for this option. Given a state $s$, a policy-over-options would select an option $\omega$ from the set of available options $\Omega$. The agent would then plan low-level actions by following its current option policy $a\sim \pi(\cdot|s,\omega)$ until the termination condition $\beta_\omega$ is satisfied. In our work, we use pre-defined skills as options and a pre-trained LLM as the policy-over-options to generate high-level options.

\subsection{LLM as a Planner}
Recent research has shown that LLMs have achieved significant success in various tasks within embodied environments \cite{wang2023survey, wang2023describe, ahn2022can}. Taking inspiration from these works, we employ a pre-trained LLM to act as a planner, generating a sequence of options using descriptions of observations and tasks. The generated plan, represented as a list of options $[\omega_k]_{k=1,...,K}$, is then executed by following the corresponding option policies. Formally, with text descriptions as input prompts, the LLM outputs a plan in the form of a sequence of options. An actor module subsequently generates low-level actions at each time step, following the option policy $\pi(a|s;\omega_k)$. The policies for the actor module, $\pi_{\omega}$, can either be hard-coded or learned from data.
\section{Related work}
LLMs have emerged as powerful tools for plan generation. There are some research works focusing on designing effective interfaces between planners and actors. In \cite{ahn2022can}, LLMs are employed to plan the entire sequence of options at the beginning of each task, enabling the agent to complete the task without further interaction with the planner. In \cite{wang2023describe}, the authors introduce a feedback system where the agent requests the LLM to generate an updated plan based on environmental feedback when the execution of the previous plan fails. This approach enhances the robustness of the acting agent in the face of environmental uncertainties. However, these methods often rely on hard-coded failure detectors or apply a threshold to limit the number of permissible MDP state-transition timesteps for an option. In \cite{ren2023robots}, a framework is proposed for measuring and aligning the uncertainty of LLM-based planners, allowing them to seek assistance from humans when necessary. In addition, \cite{dasgupta2023collaborating} introduce the Planner-Actor-Reporter framework, which includes a reporter module to enhance information exchange between the actor and the LLM-based planner. In this framework, the agent interacts with the LLM at each timestep, regardless of whether new information is acquired or not. While this approach eliminates the need for hard-coded termination conditions and reduces uncertainties during option execution, it leads to excessive resource consumption, especially when utilizing a large-scale and expensive LLM as the planner.

In this paper, we propose learning an interaction policy that enables the agent to interact with a remote LLM in an autonomous and ``smarter'' way. We empirically demonstrate that our approach overcomes the limitations of previously mentioned hard-coded rule-based interaction protocols or protocols that entail querying the LLM at each timestep.

\section{Our Approach When2Ask}
\label{sec: method}
We design \textit{When2Ask} based on the Planner-Actor-Mediator framework \cite{dasgupta2023collaborating}. In particular, we enhance this framework by incorporating an mediator model that learns to facilitate intelligent and cost-effective interactions between the agent and the LLM using RL.
\subsection{The Planner-Actor-Mediator Framework}
\label{sec: framework}
\begin{figure*}[t]
    \centering
    \includegraphics[width= \textwidth]{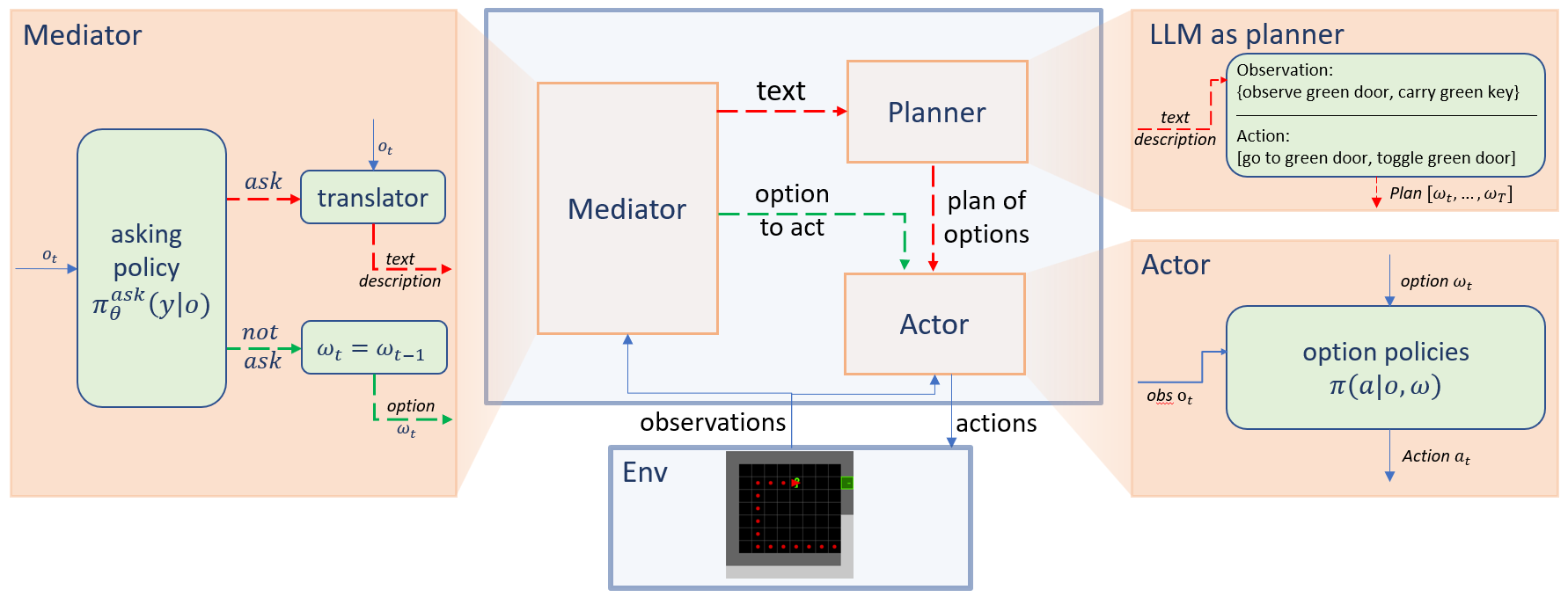}
    \caption{An overview of the Planner-Actor-Mediator paradigm and an example of the interactions. At each time step, the mediator takes the observation $o_t$ as input and decides whether to ask the LLM planner for new instructions or not. When the asking policy decides to \textcolor{red}{\textit{ask}}, as demonstrated with a \textcolor{red}{red dashed line}, the translator converts $o_t$ into text descriptions, and the planner outputs a new plan accordingly for the actor to follow. On the other hand, when the mediator decides to \textcolor{green}{\textit{not ask}}, as demonstrated with a \textcolor{green}{green dashed line}, the mediator returns to the actor directly, telling it to continue with the current plan.}
    \label{fig: framework}
\end{figure*}

This framework consists of three components, as illustrated in Fig.~\ref{fig: framework}: the planner, the actor and the mediator. The planner component is responsible for providing high-level instructions to guide the agent's actions. The actor component generates low-level actions based on these instructions. Lastly, the mediator acts as an interface between the planner and the actor, facilitating communication and coordination between them.

\keypoint{Planner} The planner component reads text-based descriptions of the current state and generates a plan for the next high-level option or a sequence of options to perform. In our framework, we utilize a pre-trained LLM as the planner. The LLM receives the descriptions of the current observation and is asked to generate high-level skill instructions for the actor. Whenever the planner is activated, the LLM generates an option plan given the descriptions provided with appropriately designed prompts.

\keypoint{Actor}
The actor component is responsible for planning the low-level actions that align with the instructed option, such as ``\textit{go to the red door}'' or ``\textit{pick up the yellow key}''. In our approach, we consider these option policies to be hard-coded using human expert knowledge. It is also possible to pre-train these policies using option-conditioned reward functions to achieve more complex skills.

\keypoint{Mediator}
In this work, our primary focus is on designing an intelligent mediator component within the Planner-Actor-Mediator framework.
Our approach involves training an explicit asking policy using RL to determine when to interact with the planner. The mediator component consists of two sub-components: an asking policy that decides whether to request a new plan from the planner based on observations and the current option, and a translator module that converts observations into text descriptions readable by the LLM. Following \cite{ahn2022can, carta2023grounding}, we assume the availability of an expert translator here. In our experiments, the translator is designed with two stages. Firstly, we extract the IDs of objects, such as keys, doors, and boxes, observed within the field of view of the agent using the built-in interface of the simulation platform. Next, we input this information into our predefined prompt template and output it to LLM in a fixed format. An example of the format can be seen in the green box of Fig.\ref{fig: prompt_example}.
The translator can also be learned from data \cite{wang2023describe, dasgupta2023collaborating}.
\subsection{Learning asking policy with RL}
Here we introduce our proposed approach to learn an asking policy for use in the mediator component.

As mentioned earlier, interacting with the LLM can be costly. Ideally, the asking policy should be trained to enable the agent to request a new plan from the LLM only when it discovers new and informative observations. The expectation is that the LLM will provide a different plan in response to these new observations. To address this, we formulate the problem as an MDP, where the state includes information about the agent's observation and current option in action. The action space consists of two actions: ``\textit{Ask}'' and ``\textit{Not Ask}''.
In this formulation, the LLM planner is considered as part of the environment that can influence state transitions. The reward function consists of both the task-related return, denoted as $r$, and an additional penalty term that penalizes unnecessary interactions. Specifically, when the asking policy decides to ask the LLM for a new plan, but the plan provided by the LLM remains the same as the current one, the agent incurs a penalty. This penalty encourages the asking policy to avoid unnecessary interactions and ensures that requesting a new plan is primarily motivated by the discovery of new informative observations.

Denote the asking policy as $\pi^{\text{ask}}$ with its parameters represented by $\theta$. We train this policy using standard on-policy RL methods, specifically Proximal Policy Optimization (PPO) \cite{schulman2017proximal}. The objective function for training the asking policy is defined as follows:
\begin{equation}
    \max_{\theta}  \sum_{t=1} \big[\gamma^t r_t - \lambda \mathbbm{1}(y_t == \textit{Ask} \land \omega_t==\omega_{t-1} ) \big],
\end{equation}
where $y_t\in \{\textit{Ask},\textit{Not Ask}\}$ represents the decision made by the asking policy at timestep $t$, $r_t$ denotes the task reward obtained at $t$, and $\omega_t$ is the planned option provided by the LLM at $t$. The penalty factor $\lambda$ is used to balance the importance of avoiding unnecessary interactions.
Note that if the decision made by the asking policy is ``Not Ask'' ($y_t == \textit{Not Ask}$), we set $\omega_t$ to be the plan executed at the previous timestep, namely let $\omega_t = \omega_{t-1}$. This ensures that if the agent decides not to ask for a new plan, it continues executing the same plan as before. During each iteration, data is collected on-policy using the model $\pi^{\text{ask}}_\theta$.
\section{Experiments}
\label{sec: experiments}
We seek to address the following questions by experiments, : can our agent effectively reduce interaction costs while maintaining a high target task completion rate, compared with baseline methods? Can our agent proactively seek assistance from an LLM in exploratory environments?
The experimental results indicate that the answer to both questions is yes. As a byproduct, we find that our approach is able to tolerate the imperfection of a crucial component, the translator in the mediator module, which is used to convert observed images into textual descriptions (refer details to the Appendix). We employ two versions of the Vicuna model (Vicuna-7b and Vicuna-13b) \cite{touvron2023llama} as LLM planners.

\subsection{Baselines}
\label{sec:baseline}
In our experiments, we considered four baseline interaction methods as follows:

\keypoint{Hard-coded}
The timing and conditions for requesting new instructions from LLMs are manually determined by human experts for each option \cite{wang2023describe}. The agent will only request a new plan from the LLM planner when specific termination conditions for the option are met. These conditions involve a goal-finishing detector and a constraint on the maximum number of allowed timesteps. For example, let's consider the option ``go to the red door.'' The termination condition for this option specifies that the agent should reach the target door location or exceed 100 timesteps spent on this option.

\keypoint{Always}
The agent queries the LLM planner at every timestep, ensuring that any newly acquired information is immediately relayed to the planner \cite{dasgupta2023collaborating}. This strategy theoretically leads to better task performance as there is no delay between gathering new information and requesting a re-plan. However, it comes with the drawback of consuming significantly more interaction resources.

\keypoint{Random} At each timestep, the agent has a fixed probability of 50\% to query the LLM for instructions.

\keypoint{Never} The agent never interacts with the LLM. Instead, the policy-over-options (i.e., the planner) is learned using RL techniques based on data collected during interactions with the environment \cite{sutton1999between, precup2000temporal}. This means that the agent learns to make decisions and generate plans without actively querying the LLM in real-time decision-making. By comparing this method with other approaches, we can assess the contribution of using an LLM as the planner. This comparison helps evaluate the effectiveness and advantages of incorporating a pre-trained LLM into the planning process.

\subsection{MiniGrid Experiments}
\label{sec:minigrid}
The MiniGrid environment consists of a collection of 2D grid-world environments with goal-oriented tasks \cite{minigrid}. In these environments, the agent must navigate within a 2D grid room and interact with specific objects to complete various tasks, such as ``open the red door'' or ``put the green ball next to the yellow box''.

\begin{figure*}[t]
    \centering
    \begin{tikzpicture}[
squarednode1/.style={rounded corners, draw=green!60, fill=green!5, thick,  text width=0.15\textwidth, minimum height=1.3cm},
squarednode2/.style={rectangle, draw=blue!10, fill=blue!5, ultra thin},
squarednode3/.style={rectangle, draw=orange!60, fill=orange!5, thick, text width=0.15\textwidth, minimum height=1.3cm}
]
\node[squarednode2]      (frame0)                              {\includegraphics[width=0.15\textwidth]{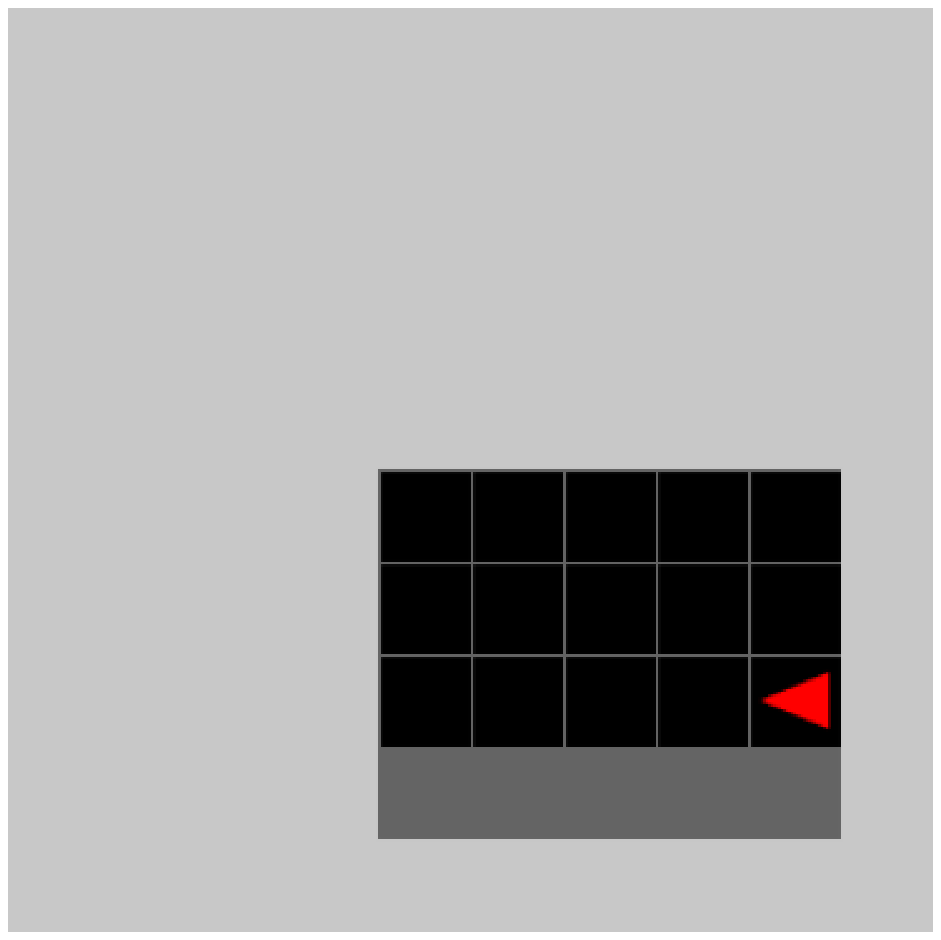}};
\node[squarednode2]        (frame1)       [right=0.4cm of frame0] {\includegraphics[width=0.15\textwidth]{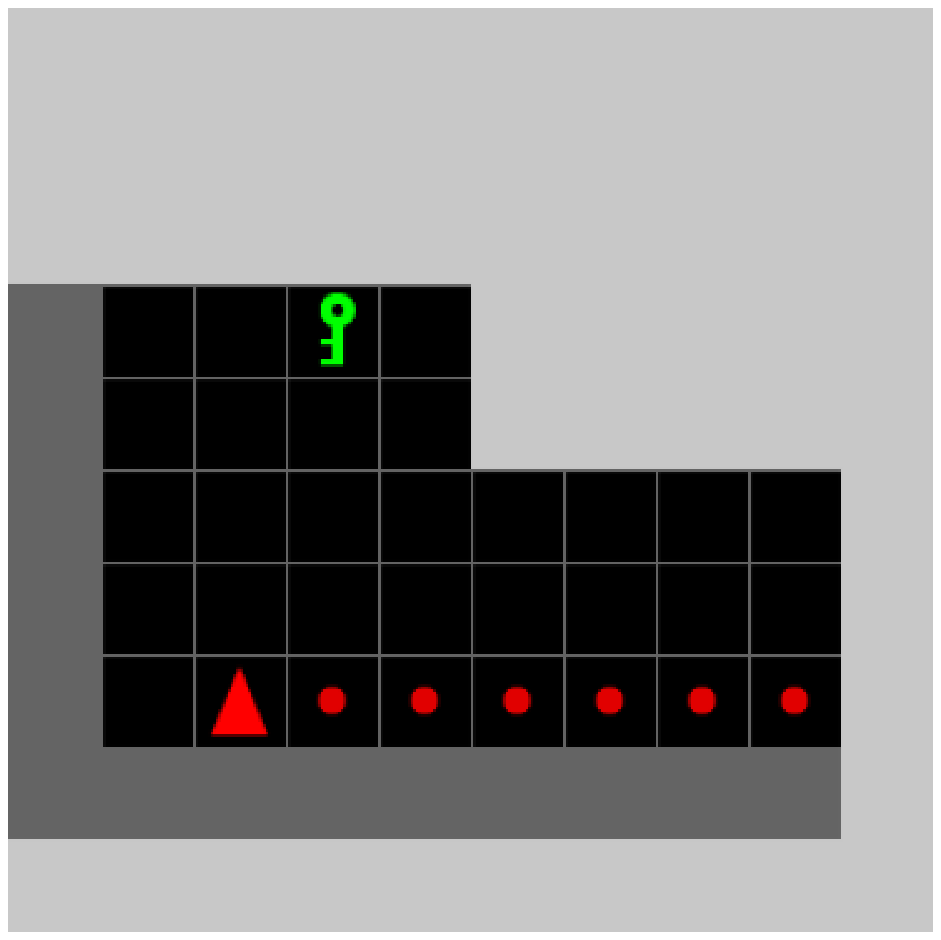}};
\node[squarednode2]      (frame2)       [right=0.4cm of frame1] {\includegraphics[width=0.15\textwidth]{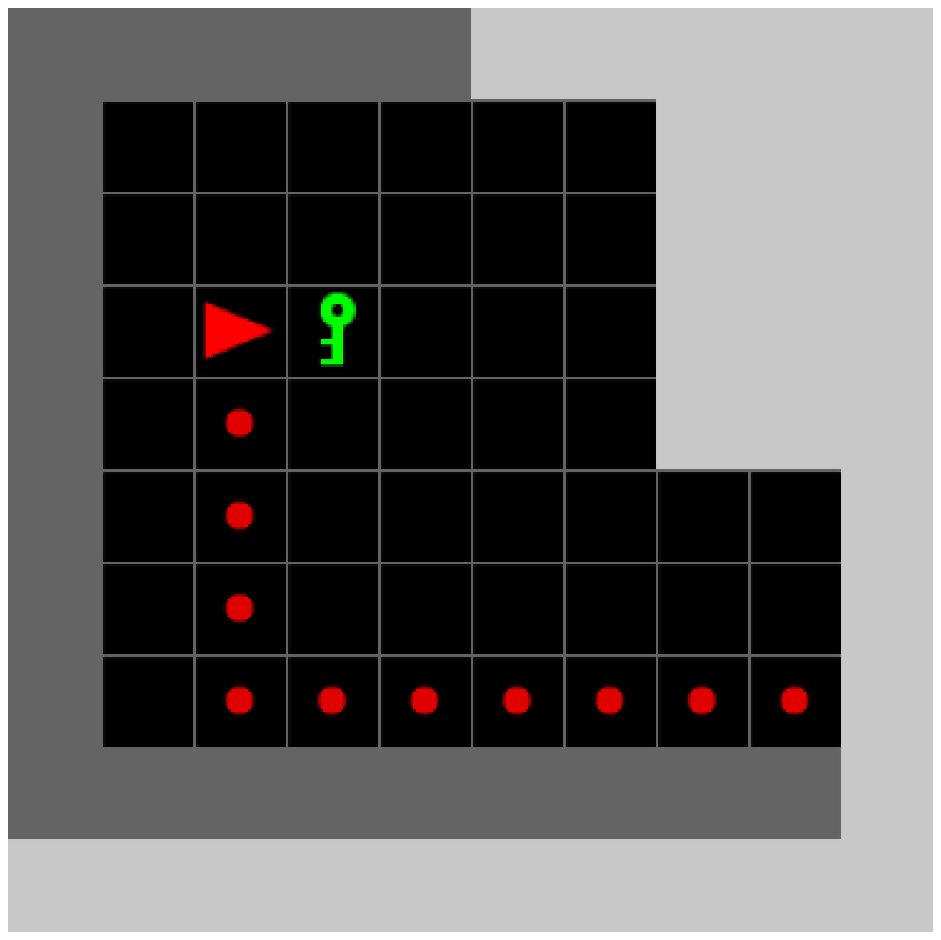}};
\node[squarednode2]        (frame3)       [right=0.4cm of frame2] {\includegraphics[width=0.15\textwidth]{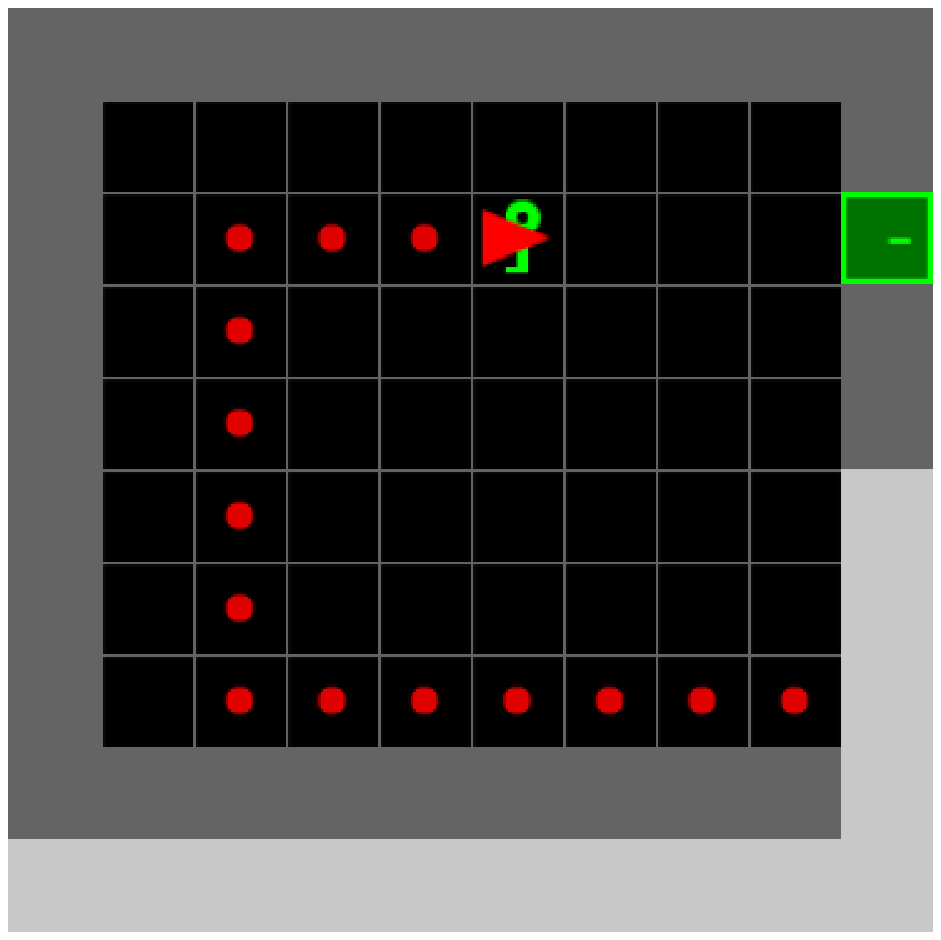}};
\node[squarednode2]        (frame4)       [right=0.4cm of frame3] {\includegraphics[width=0.15\textwidth]{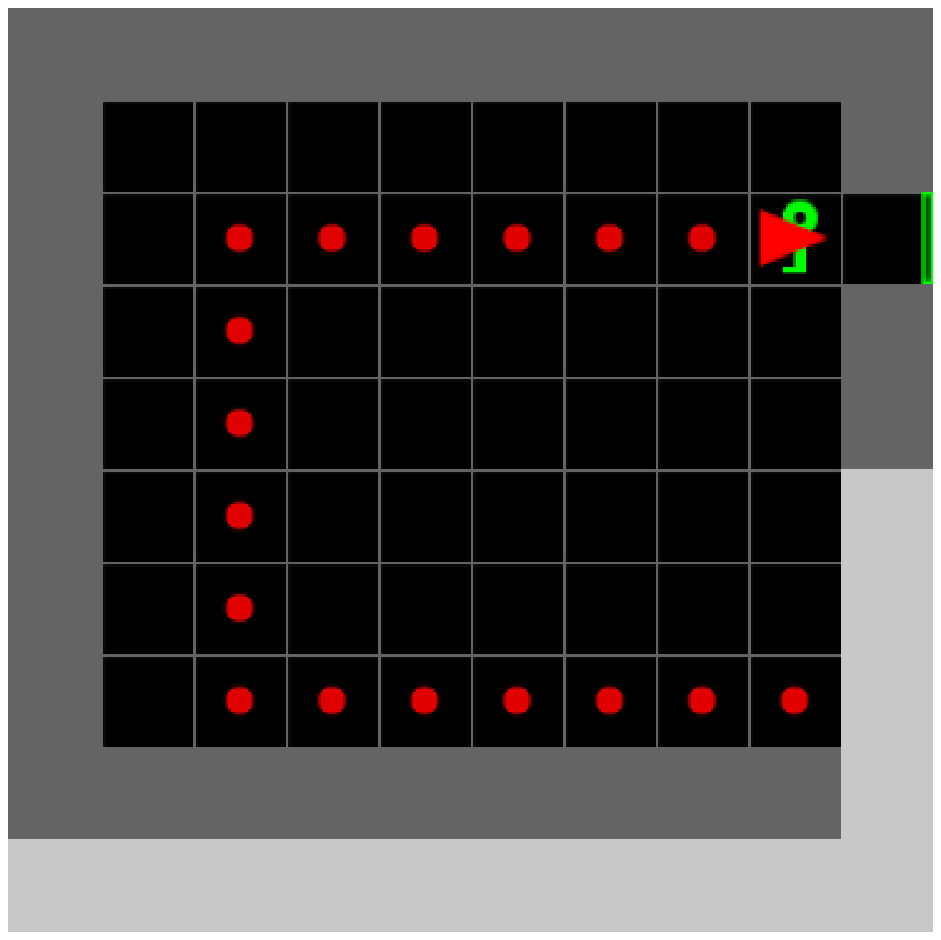}};
\node[squarednode1]    (text0) [below=0.5cm of frame0]
{\small{observed nothing}};
\node[squarednode1]    (text1) [below=0.5cm of frame1]
{\small{observed green key}};
\node[squarednode1]    (text2) [below=0.5cm of frame2]
{\small{observed green key}};
\node[squarednode1]    (text3) [below=0.5cm of frame3]
{\small{observed green door, carrying green key}};
\node[squarednode3]    (text4) [below=0.5cm of frame4]
{\small{Finished!}};
\draw[->] (frame0.east) -- (frame1.west);
\draw[->] (frame1.east) -- (frame2.west);
\draw[->] (frame2.east) -- (frame3.west);
\draw[->] (frame3.east) -- (frame4.west);

\draw[->, dashed] (frame0.south) -- (text0.north);
\draw[->, dashed] (frame1.south) -- (text1.north);
\draw[->, dashed] (frame2.south) -- (text2.north);
\draw[->, dashed] (frame3.south) -- (text3.north);

\end{tikzpicture}
    \caption{An illustrative example of the partial observations and their corresponding text descriptions in environment \textit{SimpleDoorKey}. The agent is illustrated with a red triangle, and the path it takes is illustrated with red dots. At the start of each episode, the agent is provided with only limited information, with the unexplored area masked (light grey). As the agent progresses in this room, it reveals more information about the room layout for the planner, until it successfully opens the locked door.}
    \label{fig: env_demo}
\end{figure*}

One important characteristic of this environment is that the agent's view range is limited. This means that the agent needs to explore the environment and gather useful information to plan its actions effectively. The environment returns observations in the form of a full grid, but with unexplored areas occluded, similar to the concept of ``fog of war'' in games like StarCraft. Technically, the observation returned by the environment has a shape of $o \in \mathbb{R}^{W\times H\times 4}$, where $W$ and $H$ represent the width and height of the grid, respectively. For an unexplored grid at location $[w,h]$, the observation returns the vector $[-1,-1,-1,-1]$. For an explored grid, the corresponding 4D vector contains information about the object ID, color ID, state ID (e.g., closed or locked for a door), and the agent's direction ID (indicating the agent's orientation if it is present at this location, or 4 otherwise). This design allows us to focus on the agent's reasoning ability and exclude potential influences from factors like memorization. Fig.~\ref{fig: env_demo} provides an example of the environment setup in the \textit{SimpleDoorKey} scenario.

\begin{figure*}[t]
    \centering
    \includegraphics[width= 0.7\textwidth]{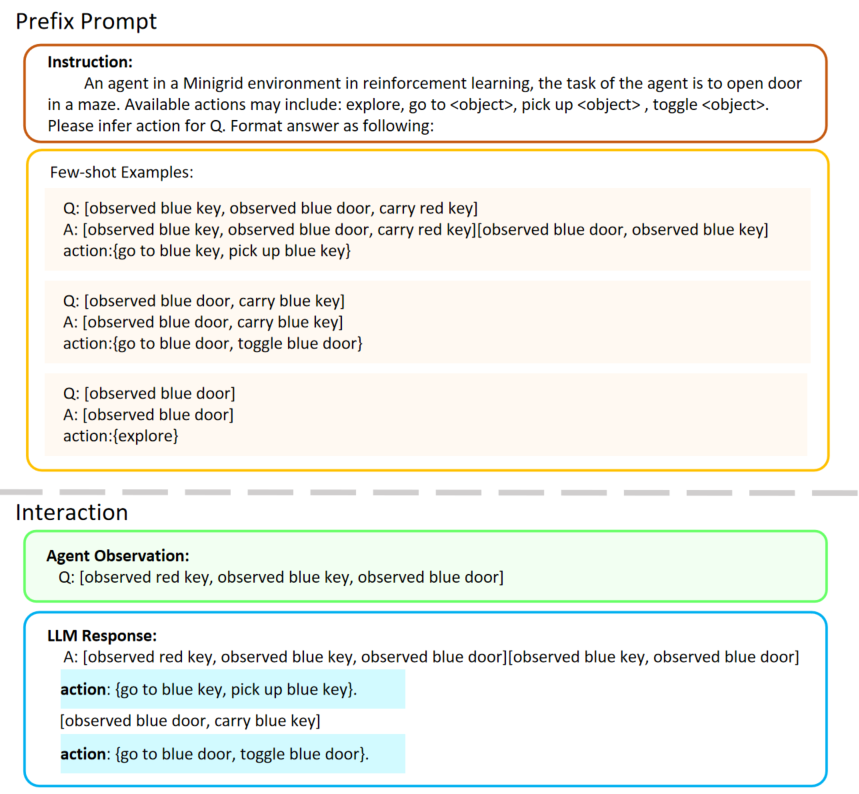}
    \caption{An example of the prefix prompt and one interaction for the \textit{ColoredDoorKey} environment. Prefix prompt consists of task instruction and few-shot examples. In Chain-of-Thought-style prompts, we add inference processes within the examples. Note that these few-shot examples are only provided for grounding a few but not all task-related knowledge to, and constraining the output formats, of the LLM. We do not need to exhaustively enumerate all knowledge and rules to construct prompts, as a qualified LLM can do logical reasoning based on a limited number of prompts, then provide proper plans (instructions) that are adaptable to new scenarios encountered in the environment.}
    \label{fig: prompt_example}
\end{figure*}
In our experiments, we focus on the task of opening a locked door in five distinct environments: \textit{SimpleDoorKey, KeyInBox, RandomBoxKey, ColoredDoorKey}, and \textit{MovingObstacle}.
All of these environments are procedurally generated, i.e., the grid layout (including room size, key and door locations) is randomly determined each time the environment is reset. To evaluate generalization, a held-out test set consisting of 100 randomly selected seeds is predefined for each environment. Refer to the Appendix for more details.

We use the Vicuna-7b model for the \textit{SimpleDoorKey}, \textit{KeyInBox}, \textit{RandomBoxKey}, and \textit{MovingObstacle} environments, while for the more complex \textit{ColoredDoorKey} environment we use the Vicuna-13b model. 
As demonstrated in previous work \cite{min2022rethinking}, language models like LLMs require carefully designed prompts and few-shot demonstrations to generalize to different tasks.
In our experiments, we provide task instructions and few-shot examples as in-context prompts for each environment. These prompts serve to guide the LLM to understand the task. For the challenging reasoning task in the \textit{ColoredDoorKey} environment, we utilize Chain-of-Thought prompts proposed by \cite{wei2022chain}. These prompts help the LLM to deal with complex reasoning tasks specific to the \textit{ColoredDoorKey} environment. The few-shot examples in the prompts are used to constraint the output formats. The LLM planner must utilize its ability to generalize and reason in order to comprehend the target task and adjust to situations that deviate from the few-shot examples, such as variations in objects' colors. Fig.~\ref{fig: prompt_example} provides an example of the prefix prompts and an interaction example in the \textit{ColoredDoorKey} environment. It shows the effective performance of the LLM planner in producing an accurate plan in response to new observations.

\subsubsection{Can our agent complete target tasks with less interaction costs?}
\begin{figure*}[t]
    \centering
    \begin{subfigure}{0.38 \textwidth}
        \includegraphics[width= \textwidth]{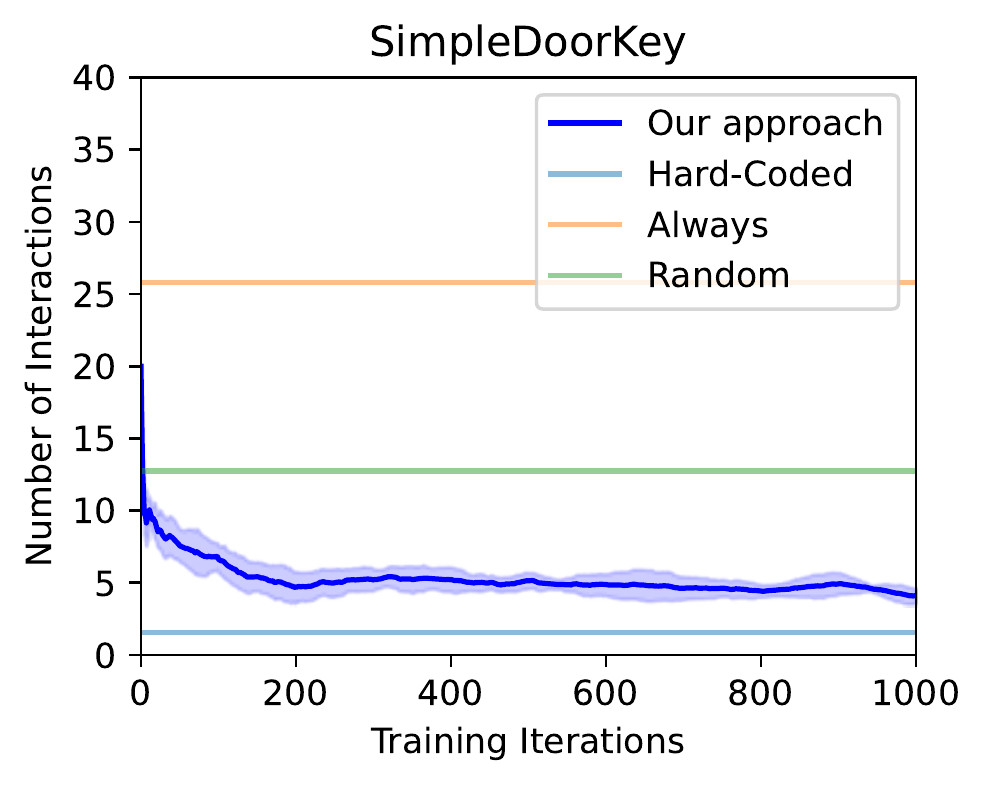}
    \end{subfigure}
    \begin{subfigure}{0.38 \textwidth}
        \includegraphics[width= \textwidth]{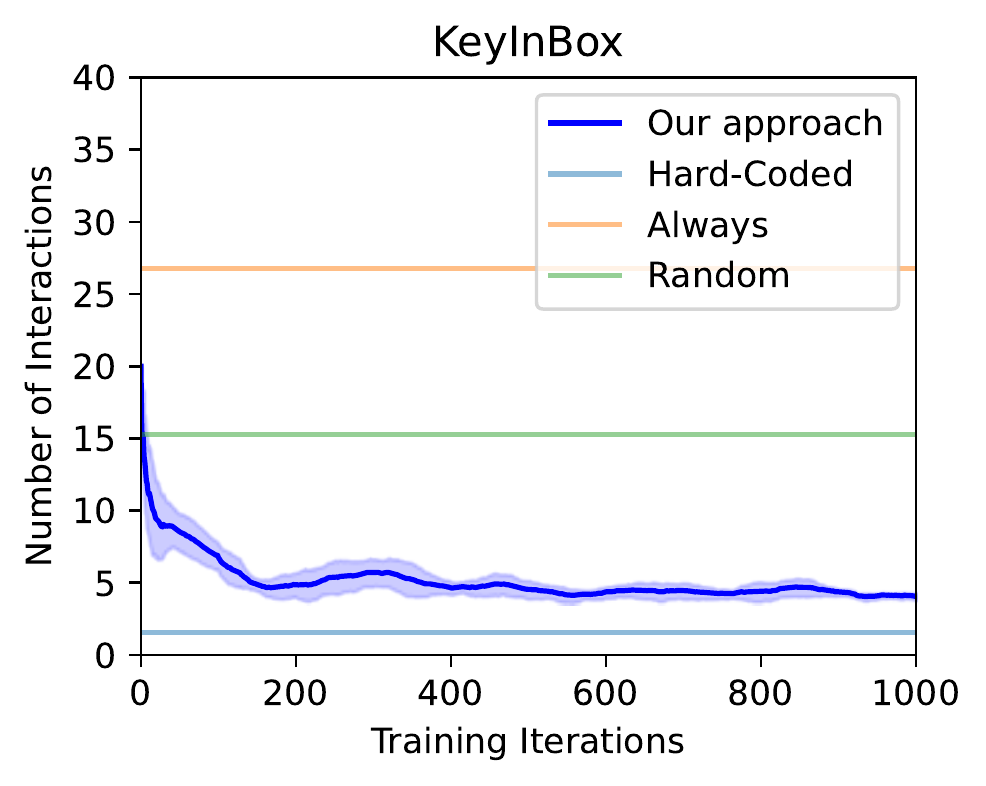}
    \end{subfigure}
    \begin{subfigure}{0.38 \textwidth}
        \includegraphics[width= \textwidth]{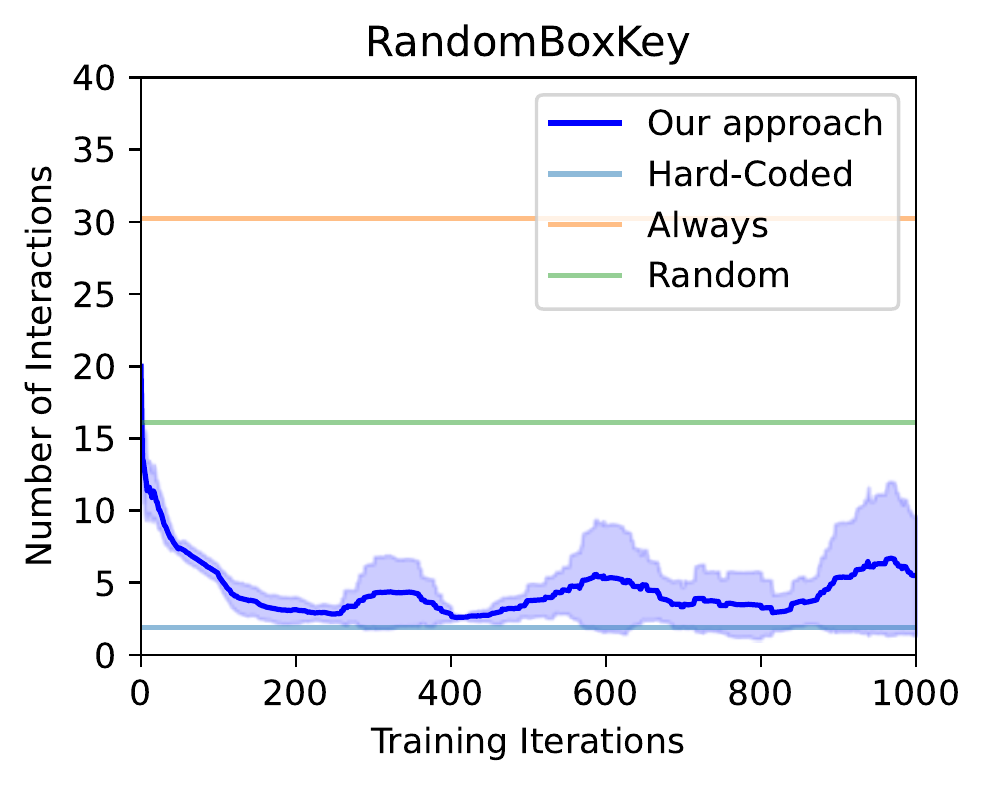}
    \end{subfigure}
    \begin{subfigure}{0.38 \textwidth}
        \includegraphics[width= \textwidth]{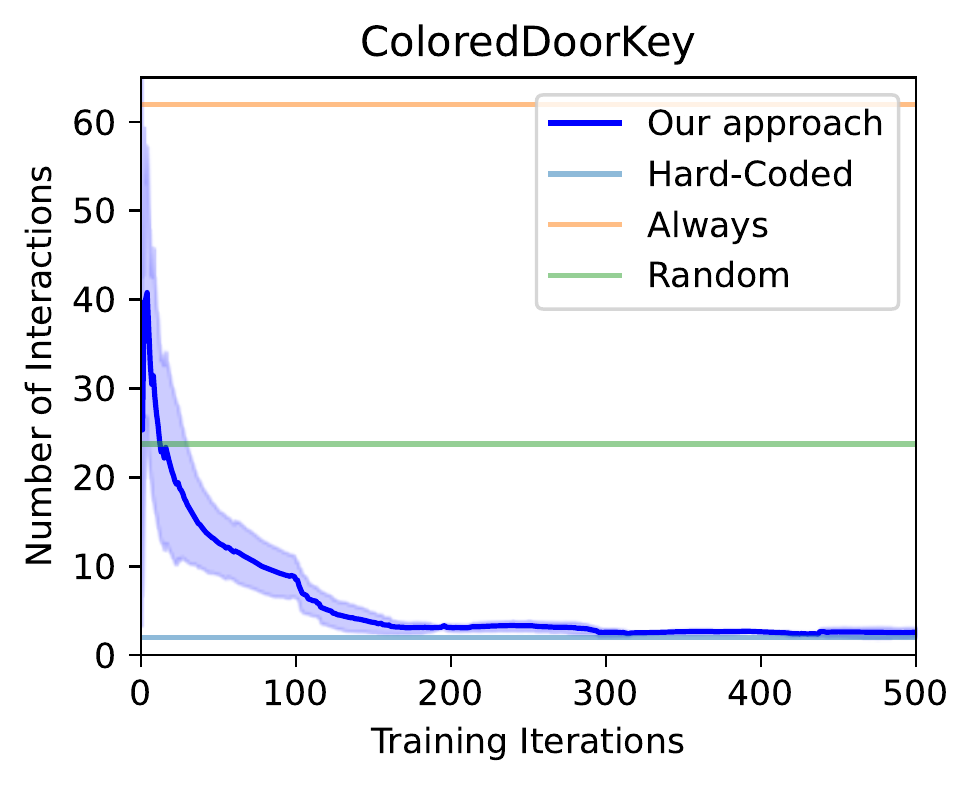}
    \end{subfigure}
     \begin{subfigure}{0.38 \textwidth}
        \includegraphics[width= \textwidth]{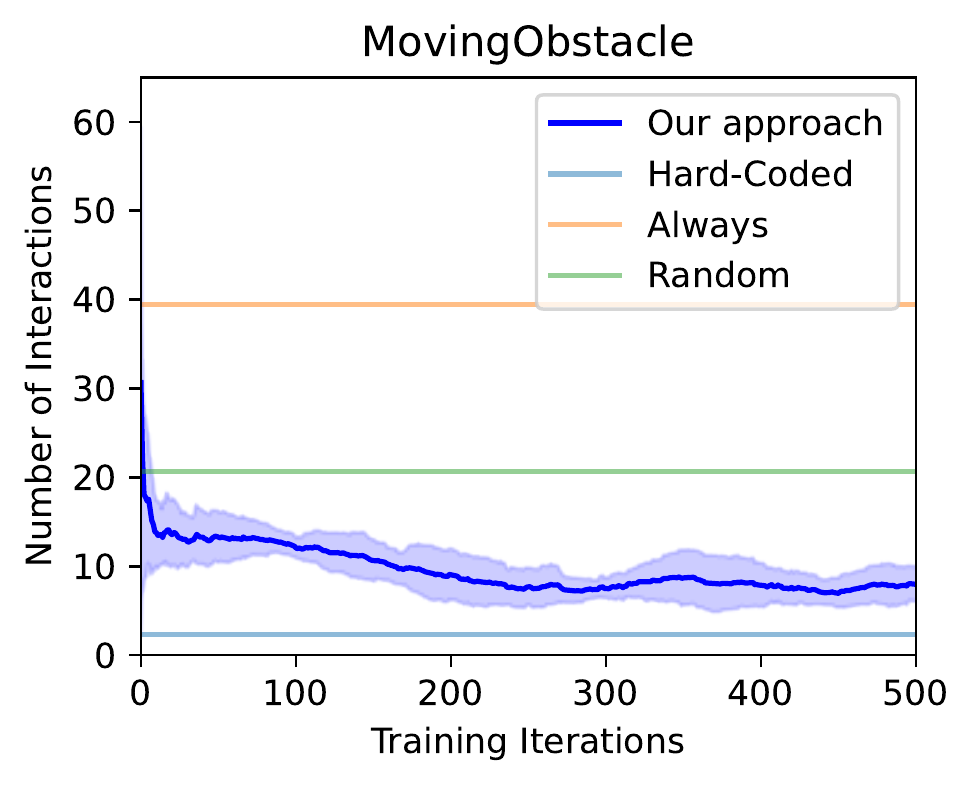}
    \end{subfigure}

    \caption{The number of interactions with the LLM vs. the number of RL iterations used for learning the asking policy. It shows that, for every environment, the more thoroughly the asking policy is trained, the fewer interactions with the LLM planner (i.e., the less interaction costs) are required to complete the task. The shaded areas within the curves represent confidence intervals based on three standard errors.}
    \label{fig: comm_curves}
\end{figure*}

\begin{figure*}[t]
    \centering
    \begin{subfigure}{0.38 \textwidth}
        \includegraphics[width= \textwidth]{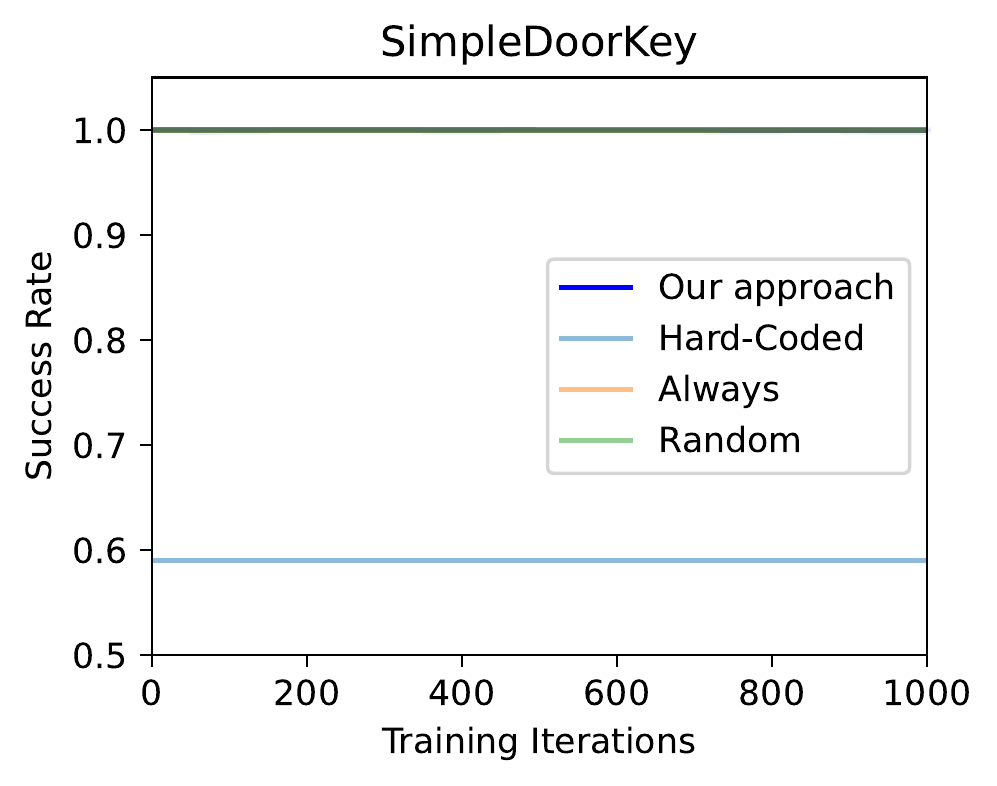}
    \end{subfigure}
    \begin{subfigure}{0.38 \textwidth}
        \includegraphics[width= \textwidth]{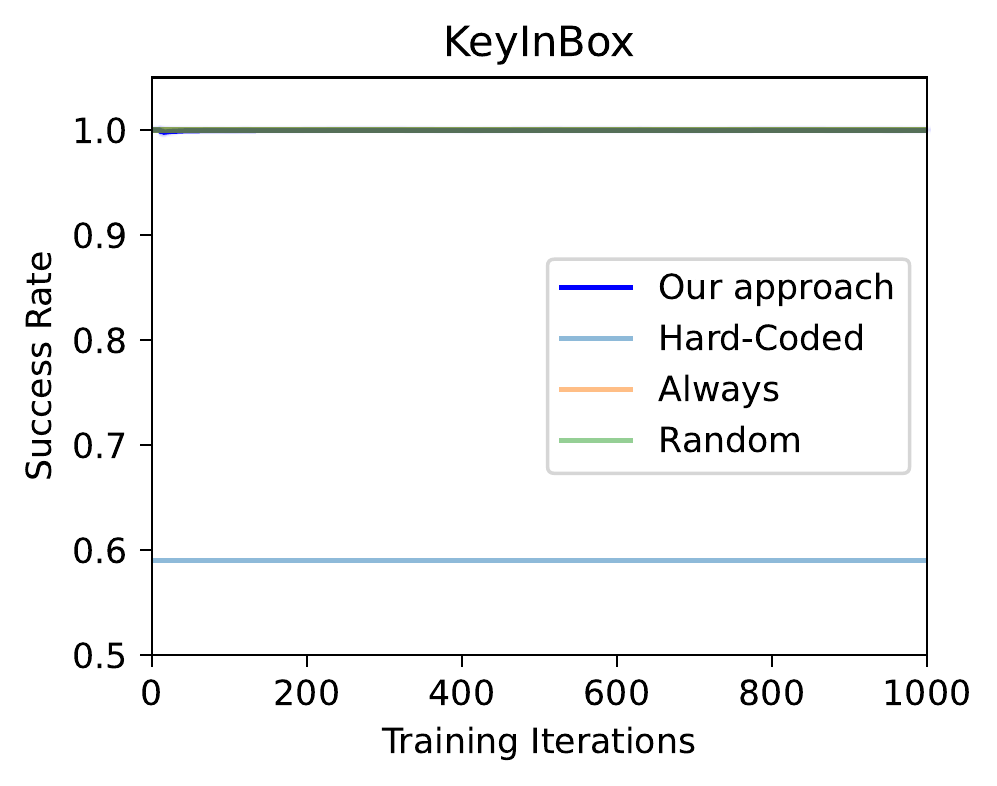}
    \end{subfigure}
    \begin{subfigure}{0.38 \textwidth}
        \includegraphics[width= \textwidth]{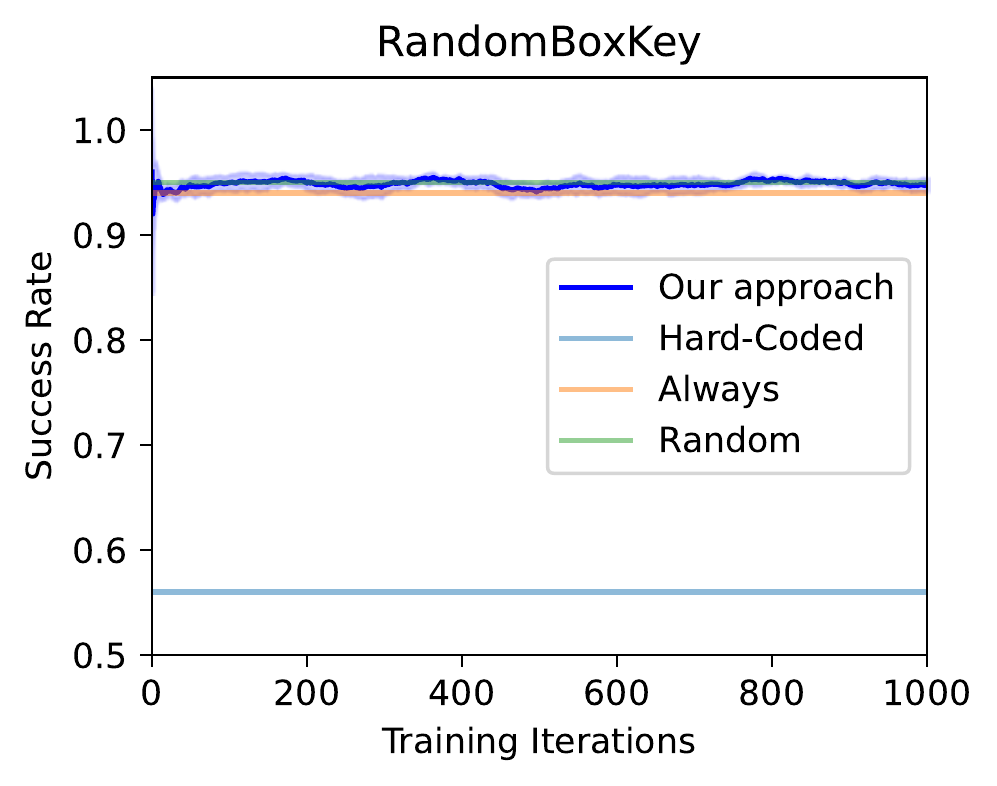}
    \end{subfigure}
    \begin{subfigure}{0.38 \textwidth}
        \includegraphics[width= \textwidth]{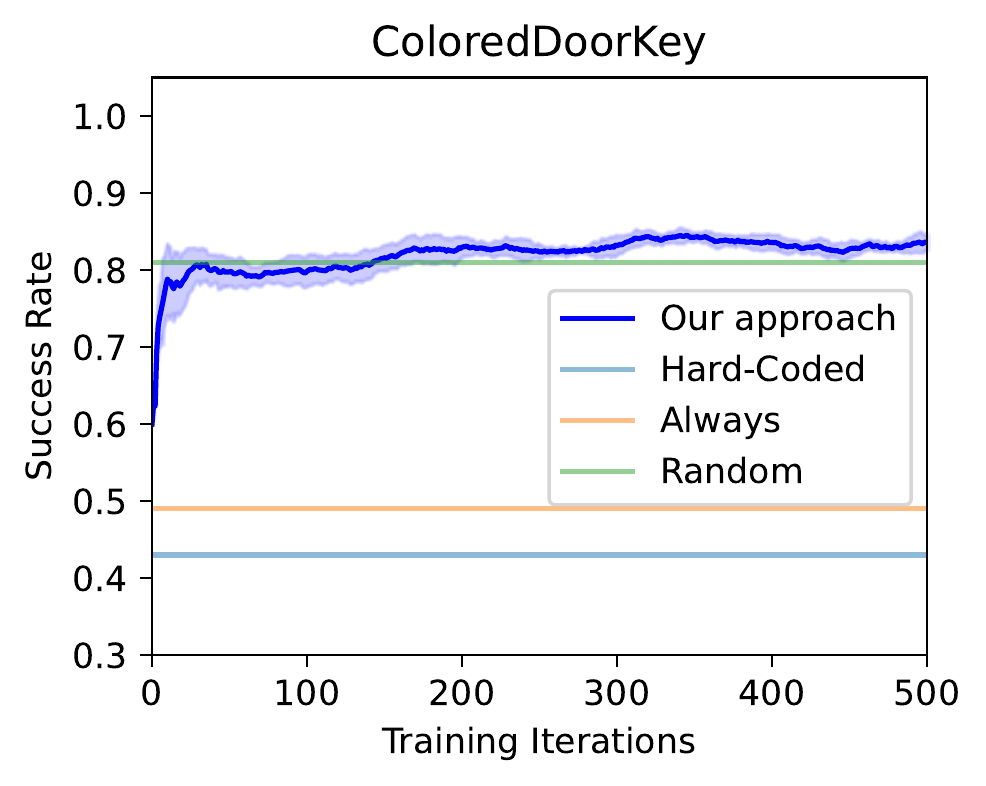}
    \end{subfigure}
    \begin{subfigure}{0.38 \textwidth}
        \includegraphics[width= \textwidth]{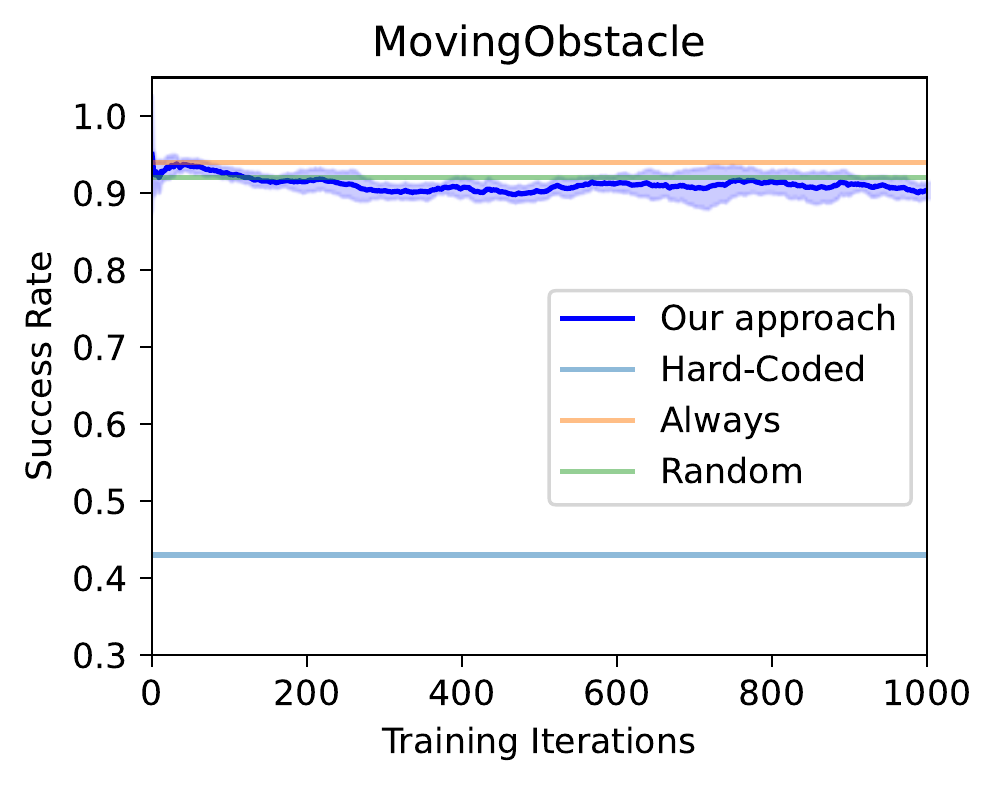}
    \end{subfigure}
    \caption{Success rate of completing target tasks vs. the number of RL iterations used for learning the asking policy. It demonstrates that our approach consistently maintains a high success rate across all environments, and outperforms baseline methods in \textit{ColoredDoorKey}.}
    \label{fig: succ_curves}
\end{figure*}

We compare our approach \textit{When2Ask} with baseline methods to evaluate its effectiveness. We analyze the learning curves for both interaction costs (Fig.~\ref{fig: comm_curves}) and task performances (Fig.~\ref{fig: succ_curves}) across all five environments. Additionally, we provide asymptotic performances in Table~\ref{tab: minigrid_result}. As is shown, our approach successfully reduces the number of interactions with the LLM while maintaining task performance across all environments. This reduction in interaction cost indicates that our method effectively learns to reduce non-informative interactions with the LLM. Furthermore, our approach maintains consistently high success rates throughout the learning process. This observation indicates that the asking policy learns to filter out unnecessary interactions while still engaging in essential ones to achieve successful task completion.
\begin{table*}[h]
\caption{Asymptotic performance comparison on five \textit{MiniGrid} environments. The performance metrics include the total number of interactions with the LLM, the number of MDP state-transition timesteps, and the success rate for completing a task. \textit{These results show that our approach achieves competitive task performance in terms of success rate while significantly reducing interaction costs (indicated by the number of interactions) compared to \textit{Always} and \textit{Random}}. \textit{Hard-coded} requires the fewest LLM interactions but often fails to complete tasks. All results are averaged over 500 test trials (We use 5 training seeds to initialize the policy network, and conduct 100 independent tests per seed).}
    \label{tab: minigrid_result}
    
    \centering
    \begin{tabular}{l | l  c c c c}
    Environment & Performance metric  & Hard-Coded & Always & Random & Our approach \\
      \hline 
    \multirow{3}{*}{\textit{SimpleDoorKey}} 
    & \textit{Number of interactions} $\downarrow$ & \textbf{1.58} & 25.78 & 12.75 & 4.24 \\
    & \textit{Number of timesteps} $\downarrow$  & 64.9 & \textbf{25.78} & 26.55 & 29.20 \\
    & \textit{Task success rate} $\uparrow$ & 59\% & \textbf{100\%} & \textbf{100\%} & \textbf{100\%} \\ 
    \hline
    \multirow{3}{*}{\textit{KeyInBox}} 
    & \textit{Number of interactions} $\downarrow$ & \textbf{1.58} & 26.78 & 15.3 & 4.33 \\
    & \textit{Number of task timesteps} $\downarrow$ & 65.49 & \textbf{26.78} & 27.46 & 29.01 \\
    & \textit{Task success rate} $\uparrow$ & 59\% &\textbf{ 100\%} & \textbf{100\%} &\textbf{ 100\%} \\
    \hline
    \multirow{3}{*}{\textit{RandomBoxKey}} 
    & \textit{Number of interactions} $\downarrow$ & \textbf{1.93} & 30.26 & 16.09 & 3.61 \\
    & \textit{Number of task timesteps} $\downarrow$ & 61.71 & 30.26 & \textbf{30.2} & 34.41 \\
    & \textit{Task success rate} $\uparrow$ & 56\% & 94\% & \textbf{95\%} & \textbf{95\%} \\ 
    \hline 
    \multirow{3}{*}{\textit{ColoredDoorKey}} 
    & \textit{Number of interactions} $\downarrow$ & \textbf{2.01} & 61.96 & 23.75 & 3.29 \\
    & \textit{Number of timesteps} $\downarrow$ & 75.54 & 61.96 & \textbf{44.64} & 47.87 \\
    & \textit{Task success rate} $\uparrow$ & 43\% & 49\% & 81\% & \textbf{83\%} \\ 
    \hline
    \multirow{3}{*}{\textit{MovingObstacle}} 
    & \textit{Number of interactions} $\downarrow$ & \textbf{2.29} & 39.49 & 20.70 & 6.94 \\
    & \textit{Number of timesteps} $\downarrow$ & 82.36 & \textbf{39.49} & 44.90 & 48.63 \\
    & \textit{Task success rate} $\uparrow$ & 43\% & \textbf{94\%} & 93\% & 92\% \\ 
    \end{tabular}
    
\end{table*}
\subsubsection{Can our agent proactively seek assistance from an LLM in exploratory environments?}
Upon analyzing the agent's performance in situations where it is expected to ask the LLM planner for help, we observe that the baseline method with a hard-coded asking policy exhibited significantly lower success rates compared to other approaches. This discrepancy occurs because the agent continues executing every option until its termination condition is met, even when it has already gathered sufficient information to complete the task. Consequently, this inefficient approach results in wasted time on each option and ultimately leads to failure in completing the task within the given time limit.
In contrast, \textit{When2Ask}, along with other baseline methods, demonstrates the ability to early-stop options when necessary. As a result, they achieve 100 percent success rates in \textit{SimpleDoorKey} and \textit{KeyInBox}.

\begin{figure}
    \centering
    \begin{subfigure}{0.45\textwidth}
        \includegraphics[width = \textwidth]{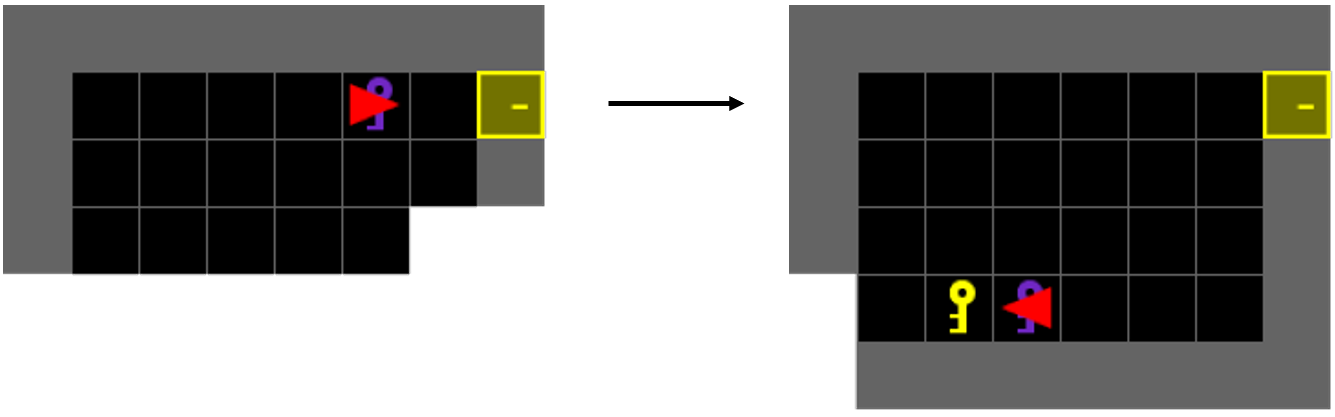}
        \subcaption{An example scenario where the agent discovers new information during option \textit{explore}.}
        \label{fig: example1}
    \end{subfigure}

    \begin{subfigure}{0.45\textwidth}
        \includegraphics[width = \textwidth]{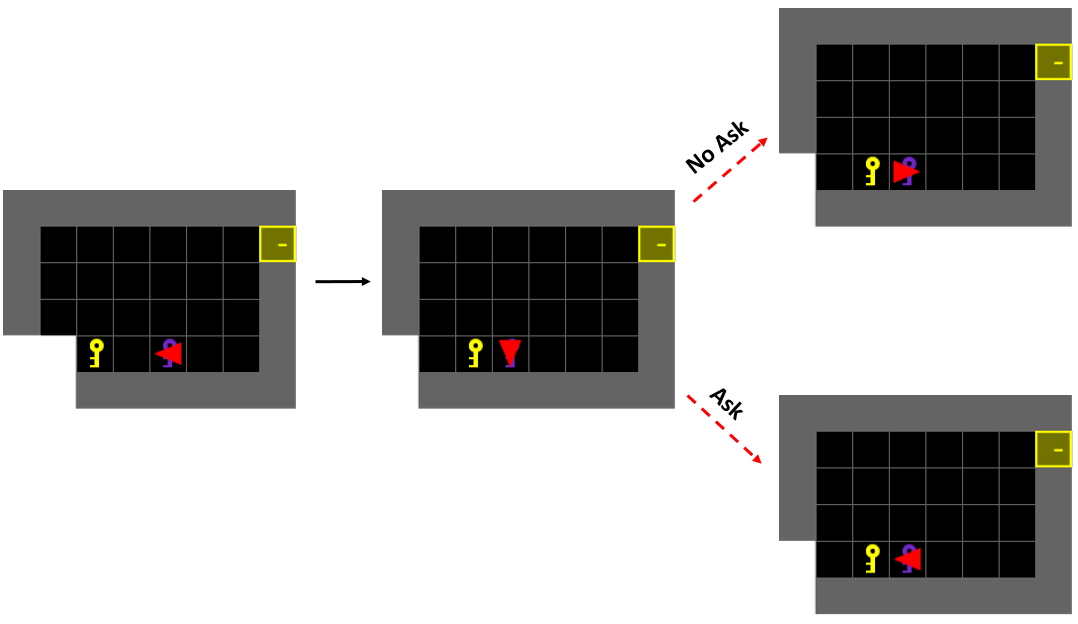}
        \subcaption{An example scenario where the hard-coded translator fails to encode all information.}
        \label{fig: example2}
    \end{subfigure}

    \caption{Two example scenarios where the agent is expected: (a) to ask the LLM planner for help as it has collected useful information for the planner to adjust its plan; and (b) to not ask the LLM, as the LLM may propose wrong options due to an imperfect translator.}
    \label{fig:examples}
    \vspace{-0.35cm}
\end{figure}

In a specific scenario within the \textit{ColoredDoorKey} environment, as illustrated in Fig.~\ref{fig: example1}, we see an interesting phenomenon. The agent has chosen to take the \textit{Explore} option and acquired information about the location of the yellow key (frame 2). With use of the \textit{Hard-coded} baseline approach, the agent shall continue with the \textit{Explore} option until it has fully explored the entire room. In contrast, using our proposed approach, the agent can recognize the value of asking the LLM planner for guidance given the current information, and immediately propose querying the LLM planner for an updated plan while ceasing further exploration. The LLM would instruct the agent to efficiently pick up the yellow key without wasting additional time.
This example highlights the effectiveness of our approach in recognizing when to seek assistance from the LLM planner and making more efficient decisions based on the available information. By leveraging the power of the LLM planner, our approach enables the agent to make informed choices that expedite task completion and improve overall performance.
\begin{figure}[t]
    \centering
    \begin{subfigure}{0.45 \textwidth}
        \includegraphics[width= \textwidth]{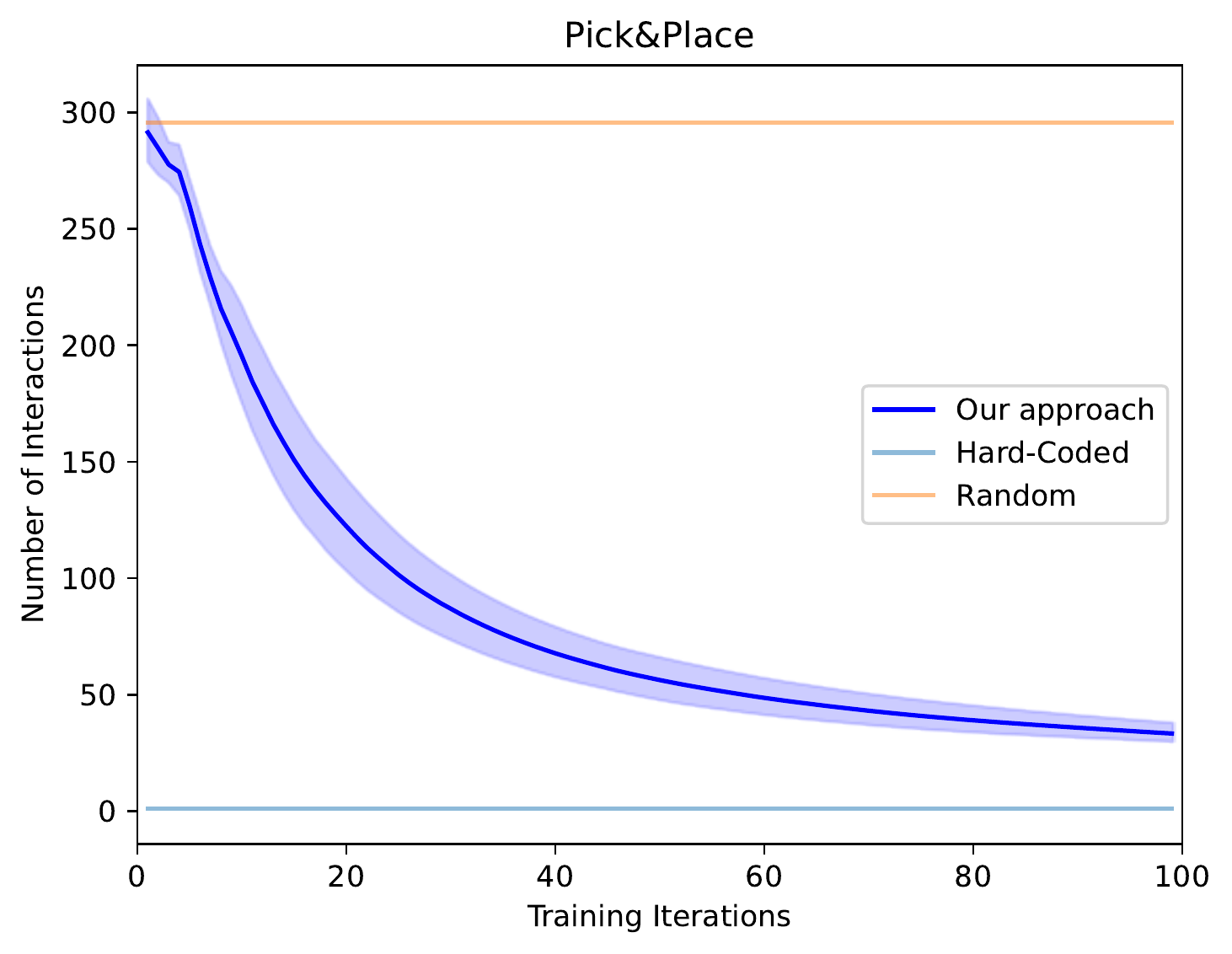}
    \end{subfigure}
    \begin{subfigure}{0.45 \textwidth}
        \includegraphics[width= \textwidth]{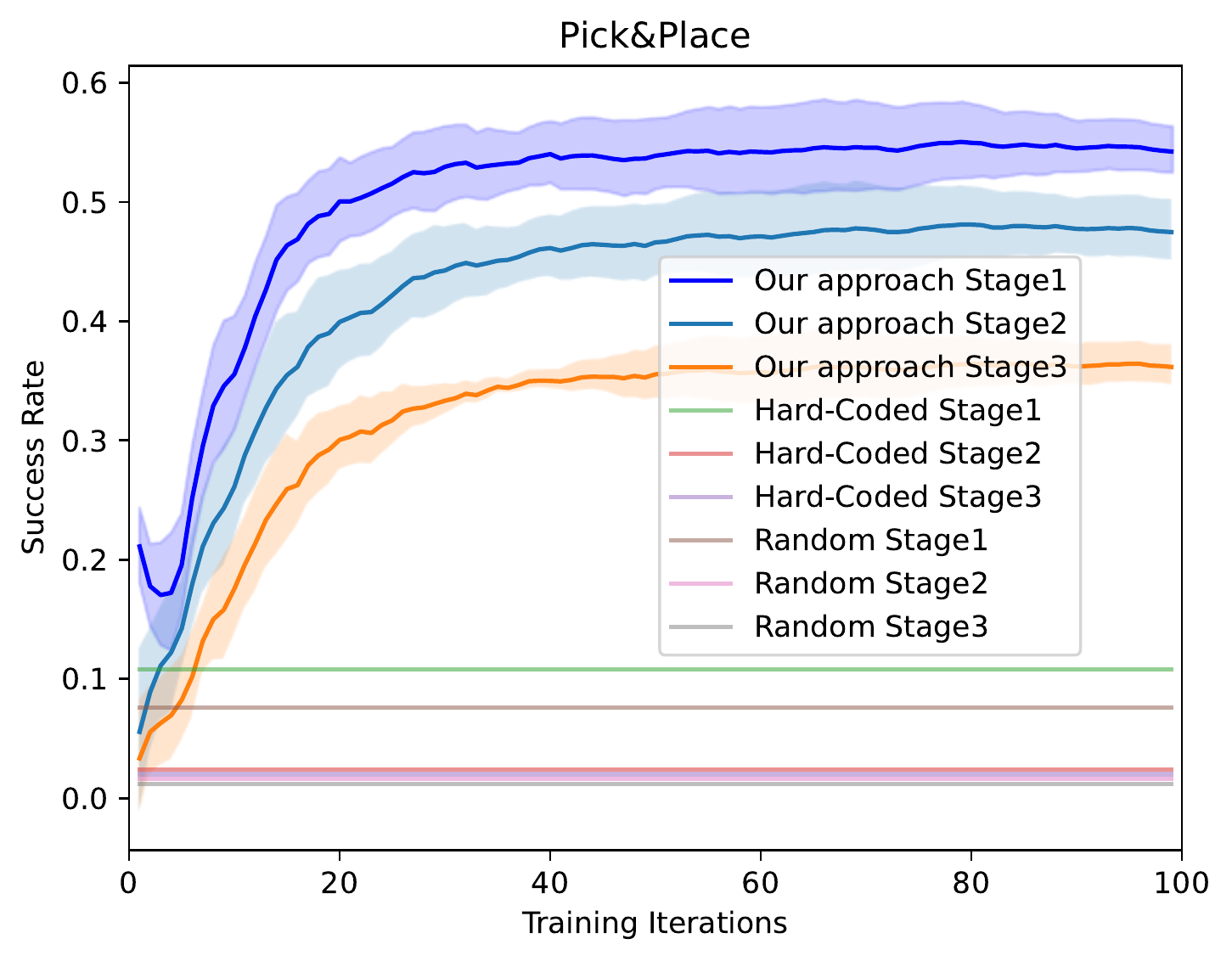}
    \end{subfigure}
    \caption{The number of interactions with the LLM (left) and the stage success rates (right) vs. the number of training iterations used for learning the asking policy on the \textit{Pick\&Place} task.}
    \label{fig: habitat}
\end{figure}
\begin{figure}
    \centering
    \begin{subfigure}{0.20 \textwidth}
        \includegraphics[width= \textwidth]{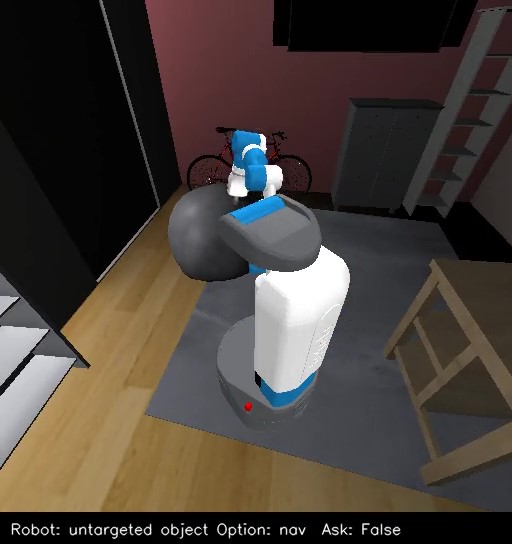}
    \end{subfigure}
    \begin{subfigure}{0.20 \textwidth}
        \includegraphics[width= \textwidth]{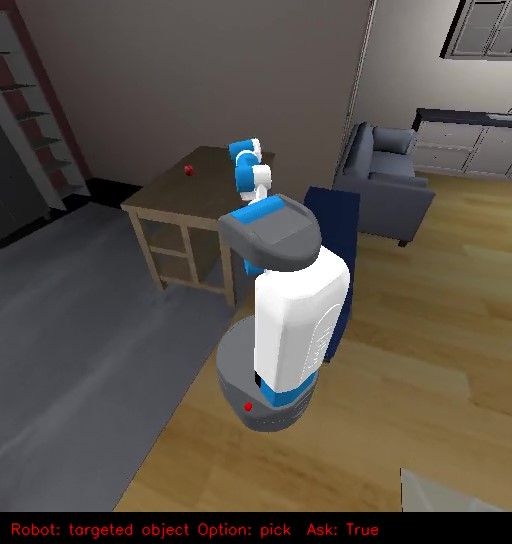}
    \end{subfigure}
    \caption{An illustrative example demonstrating the “hand-off” problem in Habitat. The robot’s objective is to navigate to the living room and pick up the apple from the table. With the \textit{Hard-Coded} baseline in use (left), according to preset hard-coded rules, the agent must first complete the \textit{Navigate} option before executing the \textit{Pick} option. Consequently, the agent stops at a location where the apple is not visible at the end of \textit{Navigate}, resulting in its future failure in the \textit{Pick} option. With our approach (right), in the middle of \textit{Navigate}, the agent finds itself at a suitable location where the apple can be spotted. The learned mediator interrupts the ongoing \textit{Navigate} and query the LLM planner, which returns the \textit{Pick} option. This helps the agent subsequently pick up the apple successfully. This example demonstrates the effectiveness of our approach in bypassing the “hand-off” issue.}
    \label{fig: habitat_skill}
\end{figure}
\subsection{Habitat Experiments}
\label{sec:habitat}
We further evaluate our approach with the Habitat environment \cite{szot2021habitat}. The results indicate the potential of our approach to function effectively in visually realistic domains. The details on the experiment setting are referred to the Appendix.
\begin{table}[t]
    \centering
        \caption{Success rate of each stage completions and total number of interactions with the LLM planner in the \textit{Habitat} during testing. }
    \begin{tabular}{c| c c c}
     Performance metric    &   Hard-Coded & Random & Our approach\\
   \hline
     \textit{Stage1 success rate}$\uparrow$  & 10.8\%& 7.6\% & \textbf{53.6\%}\\
     \textit{Stage2 success rate}$\uparrow$  & 2.4\% & 1.6\% & \textbf{46.4\%}\\
     \textit{Stage3 success rate}$\uparrow$  & 2.0\% & 1.2\% & \textbf{35.6\%} \\
     \textit{Total \# of interactions}$\downarrow$ &
     \textbf{1.00} & 295.60 & 7.99
    \end{tabular}
    \label{tab: habitat}
\end{table}

We compare our approach against baselines on the \textit{Pick\&Place} task. To ensure reliability of experimental results, we utilize 10 training seeds to initialize the policy network. This allows us to explore different initializations and avoid biased results. Subsequently, we select the best policy obtained from these training runs to run 250 independent testing trials.
As presented in Table~\ref{tab: habitat} and Fig.~\ref{fig: habitat}, our approach significantly outperforms baselines across all three stages. Particularly, compared to the hard-coded baseline where the preset plan is executed step-by-step, our approach is significantly better at addressing the ``hand-off problem'' \cite{szot2021habitat} that can arise when the preceding option terminates at a state that makes it challenging for the succeeding option to initiate. This issue is depicted in Fig.~\ref{fig: habitat_skill}, where the robot stops at an unfavorable location at the end of the \textit{Navigate} option, resulting in a failure to execute the subsequent \textit{Pick} option. Our approach effectively bypasses this problem by seeking guidance from the LLM planner.

The obtained results demonstrate that the RL learned asking policy effectively establishes a connection between the world knowledge embedded within the LLMs and the downstream fine-grained knowledge embedded within the agent. This connection leads to a superior overall performance of our approach compared to the baselines that do not involve any learning. These findings align with the main observations from our experiments associated with the MiniGrid environments, particularly in the \textit{ColoredDoorKey} scenario, where the RL learned asking policy enables the agent to outperform all baselines in terms of task completion success rate.
\section{Concluding Remarks}
\label{sec: conclusion}
We examine the application of RL in acquiring ``mediator'' policies for instruction-following agents powered by LLMs. Prior research has indicated that LLMs, when coupled with well-constructed prompts, can effectively generate high-level instructions conditional on state descriptions to devise detailed plans for task completion. Recent frameworks for LLM-driven planning have explored two primary strategies: 1) generating an updated plan at each timestep, and 2) requesting an update only after each plan (option) concludes based on predefined termination criteria. The former method is computationally intensive due to the expense of generating a new response from the LLM, while the latter may encounter challenges as an ongoing plan executed cannot be interrupted in time to respond to new observations.

To this end, we propose \textit{When2Ask}, which involves training a mediator policy to determine when to prompt the LLM to generate an appropriate plan for the present moment. \textit{When2Ask} trains the mediator policy to optimize task-oriented rewards while penalizing cases where the planner was invoked but returned the same plan the agent was already following. Experiment results across different embodied environments illustrate that the learned mediator policies achieve comparable task success rates to fixed policies that query the LLM at each timestep, while significantly reducing the number of queries to the LLM and consequently lowering the computational burden on the agent.

The utilization of LLMs to furnish robots and other autonomous agents with general-purpose reasoning and planning capabilities has shown great potential. However, this potential is somewhat limited by the quality of the mapping between low-level observations and actions, and the high-level LLM-based planner. A long-term goal would be to learn this mapping in an end-to-end way. In the interim, however, it is worth to investigate how different elements of this mapping can be learned, and how much benefit can be gained from such an endeavor, as demonstrated in this study.

\subsubsection*{Acknowledgments}
\label{sec:ack}
We appreciate anonymous reviewers for valuable comments that helped us to improve the quality of this paper. This work was primarily supported by Exploratory Research Project (No.2022RC0AN02) of Zhejiang Lab, and partly supported by China Postdoctoral Science Foundation (No.2023M743266), and Zhejiang Provincial Postdoctoral Research Project (No.ZJ2023067). Z. Xu was partly supported by Fudan University (No.JIH2325015Y).


\bibliography{ref}

\begin{thebibliography}{23}
\providecommand{\natexlab}[1]{#1}
\providecommand{\url}[1]{\texttt{#1}}
\expandafter\ifx\csname urlstyle\endcsname\relax
  \providecommand{\doi}[1]{doi: #1}\else
  \providecommand{\doi}{doi: \begingroup \urlstyle{rm}\Url}\fi

\bibitem[Ahn et~al.(2022)Ahn, Brohan, Brown, Chebotar, Cortes, David, Finn,
  Gopalakrishnan, Hausman, Herzog, et~al.]{ahn2022can}
Michael Ahn, Anthony Brohan, Noah Brown, Yevgen Chebotar, Omar Cortes, Byron
  David, Chelsea Finn, Keerthana Gopalakrishnan, Karol Hausman, Alex Herzog,
  et~al.
\newblock Do as i can, not as i say: Grounding language in robotic affordances.
\newblock \emph{arXiv preprint arXiv:2204.01691}, 2022.

\bibitem[Brown et~al.(2020)Brown, Mann, Ryder, Subbiah, Kaplan, Dhariwal,
  Neelakantan, Shyam, Sastry, Askell, et~al.]{brown2020language}
Tom Brown, Benjamin Mann, Nick Ryder, Melanie Subbiah, Jared~D Kaplan, Prafulla
  Dhariwal, Arvind Neelakantan, Pranav Shyam, Girish Sastry, Amanda Askell,
  et~al.
\newblock Language models are few-shot learners.
\newblock \emph{Advances in neural information processing systems},
  33:\penalty0 1877--1901, 2020.

\bibitem[Carta et~al.(2023)Carta, Romac, Wolf, Lamprier, Sigaud, and
  Oudeyer]{carta2023grounding}
Thomas Carta, Cl{\'e}ment Romac, Thomas Wolf, Sylvain Lamprier, Olivier Sigaud,
  and Pierre-Yves Oudeyer.
\newblock Grounding large language models in interactive environments with
  online reinforcement learning.
\newblock \emph{arXiv preprint arXiv:2302.02662}, 2023.

\bibitem[Chevalier-Boisvert et~al.(2018)Chevalier-Boisvert, Bahdanau, Lahlou,
  Willems, Saharia, Nguyen, and Bengio]{chevalier2018babyai}
Maxime Chevalier-Boisvert, Dzmitry Bahdanau, Salem Lahlou, Lucas Willems,
  Chitwan Saharia, Thien~Huu Nguyen, and Yoshua Bengio.
\newblock Babyai: A platform to study the sample efficiency of grounded
  language learning.
\newblock \emph{arXiv preprint arXiv:1810.08272}, 2018.

\bibitem[Chevalier-Boisvert et~al.(2023)Chevalier-Boisvert, Dai, Towers,
  de~Lazcano, Willems, Lahlou, Pal, Castro, and Terry]{minigrid}
Maxime Chevalier-Boisvert, Bolun Dai, Mark Towers, Rodrigo de~Lazcano, Lucas
  Willems, Salem Lahlou, Suman Pal, Pablo~Samuel Castro, and Jordan Terry.
\newblock Minigrid \& miniworld: Modular \& customizable reinforcement learning
  environments for goal-oriented tasks.
\newblock \emph{CoRR}, abs/2306.13831, 2023.

\bibitem[Das et~al.(2018)Das, Datta, Gkioxari, Lee, Parikh, and
  Batra]{das2018embodied}
Abhishek Das, Samyak Datta, Georgia Gkioxari, Stefan Lee, Devi Parikh, and
  Dhruv Batra.
\newblock Embodied question answering.
\newblock In \emph{Proceedings of the IEEE conference on computer vision and
  pattern recognition}, pp.\  1--10, 2018.

\bibitem[Dasgupta et~al.(2023)Dasgupta, Kaeser-Chen, Marino, Ahuja, Babayan,
  Hill, and Fergus]{dasgupta2023collaborating}
Ishita Dasgupta, Christine Kaeser-Chen, Kenneth Marino, Arun Ahuja, Sheila
  Babayan, Felix Hill, and Rob Fergus.
\newblock Collaborating with language models for embodied reasoning.
\newblock \emph{arXiv preprint arXiv:2302.00763}, 2023.

\bibitem[Deitke et~al.(2022)Deitke, Batra, Bisk, Campari, Chang, Chaplot, Chen,
  D'Arpino, Ehsani, Farhadi, et~al.]{deitke2022retrospectives}
Matt Deitke, Dhruv Batra, Yonatan Bisk, Tommaso Campari, Angel~X Chang,
  Devendra~Singh Chaplot, Changan Chen, Claudia~P{\'e}rez D'Arpino, Kiana
  Ehsani, Ali Farhadi, et~al.
\newblock Retrospectives on the embodied ai workshop.
\newblock \emph{arXiv preprint arXiv:2210.06849}, 2022.

\bibitem[Huang et~al.(2022)Huang, Xia, Xiao, Chan, Liang, Florence, Zeng,
  Tompson, Mordatch, Chebotar, et~al.]{huang2022inner}
Wenlong Huang, Fei Xia, Ted Xiao, Harris Chan, Jacky Liang, Pete Florence, Andy
  Zeng, Jonathan Tompson, Igor Mordatch, Yevgen Chebotar, et~al.
\newblock Inner monologue: Embodied reasoning through planning with language
  models.
\newblock \emph{arXiv preprint arXiv:2207.05608}, 2022.

\bibitem[Jiang et~al.(2022)Jiang, Gupta, Zhang, Wang, Dou, Chen, Fei-Fei,
  Anandkumar, Zhu, and Fan]{jiang2022vima}
Yunfan Jiang, Agrim Gupta, Zichen Zhang, Guanzhi Wang, Yongqiang Dou, Yanjun
  Chen, Li~Fei-Fei, Anima Anandkumar, Yuke Zhu, and Linxi Fan.
\newblock Vima: General robot manipulation with multimodal prompts.
\newblock \emph{arXiv preprint arXiv:2210.03094}, 2022.

\bibitem[Min et~al.(2022)Min, Lyu, Holtzman, Artetxe, Lewis, Hajishirzi, and
  Zettlemoyer]{min2022rethinking}
Sewon Min, Xinxi Lyu, Ari Holtzman, Mikel Artetxe, Mike Lewis, Hannaneh
  Hajishirzi, and Luke Zettlemoyer.
\newblock Rethinking the role of demonstrations: What makes in-context learning
  work?
\newblock \emph{arXiv preprint arXiv:2202.12837}, 2022.

\bibitem[Precup(2000)]{precup2000temporal}
Doina Precup.
\newblock \emph{Temporal abstraction in reinforcement learning}.
\newblock University of Massachusetts Amherst, 2000.

\bibitem[Radford et~al.(2019)Radford, Wu, Child, Luan, Amodei, Sutskever,
  et~al.]{radford2019language}
Alec Radford, Jeffrey Wu, Rewon Child, David Luan, Dario Amodei, Ilya
  Sutskever, et~al.
\newblock Language models are unsupervised multitask learners.
\newblock \emph{OpenAI blog}, 1\penalty0 (8):\penalty0 9, 2019.

\bibitem[Ren et~al.(2023)Ren, Dixit, Bodrova, Singh, Tu, Brown, Xu, Takayama,
  Xia, and Varley]{ren2023robots}
Allen~Z Ren, Anushri Dixit, Alexandra Bodrova, Sumeet Singh, Stephen Tu, Noah
  Brown, Peng Xu, Leila Takayama, Fei Xia, and Jake Varley.
\newblock Robots that ask for help: Uncertainty alignment for large language
  model planners.
\newblock In \emph{7th Annual Conference on Robot Learning}, 2023.

\bibitem[Schulman et~al.(2017)Schulman, Wolski, Dhariwal, Radford, and
  Klimov]{schulman2017proximal}
John Schulman, Filip Wolski, Prafulla Dhariwal, Alec Radford, and Oleg Klimov.
\newblock Proximal policy optimization algorithms.
\newblock \emph{arXiv preprint arXiv:1707.06347}, 2017.

\bibitem[Sutton et~al.(1999)Sutton, Precup, and Singh]{sutton1999between}
Richard~S Sutton, Doina Precup, and Satinder Singh.
\newblock Between mdps and semi-mdps: A framework for temporal abstraction in
  reinforcement learning.
\newblock \emph{Artificial intelligence}, 112\penalty0 (1-2):\penalty0
  181--211, 1999.

\bibitem[Szot et~al.(2021)Szot, Clegg, Undersander, Wijmans, Zhao, Turner,
  Maestre, Mukadam, Chaplot, Maksymets, et~al.]{szot2021habitat}
Andrew Szot, Alexander Clegg, Eric Undersander, Erik Wijmans, Yili Zhao, John
  Turner, Noah Maestre, Mustafa Mukadam, Devendra~Singh Chaplot, Oleksandr
  Maksymets, et~al.
\newblock Habitat 2.0: Training home assistants to rearrange their habitat.
\newblock \emph{Advances in Neural Information Processing Systems},
  34:\penalty0 251--266, 2021.

\bibitem[Touvron et~al.(2023)Touvron, Lavril, Izacard, Martinet, Lachaux,
  Lacroix, Rozi{\`e}re, Goyal, Hambro, Azhar, Rodriguez, Joulin, Grave, and
  Lample]{touvron2023llama}
Hugo Touvron, Thibaut Lavril, Gautier Izacard, Xavier Martinet, Marie-Anne
  Lachaux, Timoth{\'e}e Lacroix, Baptiste Rozi{\`e}re, Naman Goyal, Eric
  Hambro, Faisal Azhar, Aurelien Rodriguez, Armand Joulin, Edouard Grave, and
  Guillaume Lample.
\newblock Llama: Open and efficient foundation language models.
\newblock \emph{arXiv preprint arXiv:2302.13971}, 2023.

\bibitem[Wang et~al.(2023{\natexlab{a}})Wang, Xie, Jiang, Mandlekar, Xiao, Zhu,
  Fan, and Anandkumar]{wang2023voyager}
Guanzhi Wang, Yuqi Xie, Yunfan Jiang, Ajay Mandlekar, Chaowei Xiao, Yuke Zhu,
  Linxi Fan, and Anima Anandkumar.
\newblock Voyager: An open-ended embodied agent with large language models.
\newblock \emph{arXiv preprint arXiv:2305.16291}, 2023{\natexlab{a}}.

\bibitem[Wang et~al.(2023{\natexlab{b}})Wang, Ma, Feng, Zhang, Yang, Zhang,
  Chen, Tang, Chen, Lin, et~al.]{wang2023survey}
Lei Wang, Chen Ma, Xueyang Feng, Zeyu Zhang, Hao Yang, Jingsen Zhang, Zhiyuan
  Chen, Jiakai Tang, Xu~Chen, Yankai Lin, et~al.
\newblock A survey on large language model based autonomous agents.
\newblock \emph{arXiv preprint arXiv:2308.11432}, 2023{\natexlab{b}}.

\bibitem[Wang et~al.(2023{\natexlab{c}})Wang, Cai, Liu, Ma, and
  Liang]{wang2023describe}
Zihao Wang, Shaofei Cai, Anji Liu, Xiaojian Ma, and Yitao Liang.
\newblock Describe, explain, plan and select: Interactive planning with large
  language models enables open-world multi-task agents.
\newblock \emph{arXiv preprint arXiv:2302.01560}, 2023{\natexlab{c}}.

\bibitem[Wei et~al.(2022)Wei, Wang, Schuurmans, Bosma, Chi, Le, and
  Zhou]{wei2022chain}
Jason Wei, Xuezhi Wang, Dale Schuurmans, Maarten Bosma, Ed~Chi, Quoc Le, and
  Denny Zhou.
\newblock Chain of thought prompting elicits reasoning in large language
  models.
\newblock \emph{arXiv preprint arXiv:2201.11903}, 2022.

\bibitem[Zhou et~al.(2024)Zhou, Hu, Zhao, Zhang, and Liu]{zhou2024large}
Zihao Zhou, Bin Hu, Chenyang Zhao, Pu~Zhang, and Bin Liu.
\newblock Large language model as a policy teacher for training reinforcement
  learning agents.
\newblock In \emph{33rd International Joint Conference on Artificial
  Intelligence (IJCAI)}, 2024.

\end{thebibliography}
\bibliographystyle{rlc}

\appendix
\section{Appendix}
\subsection{Experimental settings on MiniGrid}
\label{sec:imp_MiniGrid}
In the basic setups of \textit{SimpleDoorKey} and \textit{KeyInBox}, each room contains only one key and one locked door. In \textit{SimpleDoorKey}, the key is placed on the floor, while in \textit{KeyInBox}, the key is inside a box. The agent needs to explore the room to locate the target door and the key/box, pick up the key, and finally use the key to unlock the target door.

In the \textit{RandomBoxKey} environment, the placement of the key is randomized, either on the floor or inside a box. The agent needs to actively plan its actions based on the feedback from the environment, adjusting its plan depending on whether it observes a key or a box.

\textit{ColoredDoorKey} introduces multiple keys and only one exit door. Each key and its corresponding door are color-coded, requiring a matching-colored key to unlock the door. This environment tests the agent's ability to identify and utilize color information for successful task completion.

\textit{MovingObstacle} adds another layer of complexity by introducing obstacles that move randomly within the room, potentially blocking the agent's path. The agent needs to navigate in this dynamically changing environment and adapt its plans accordingly based on new observations.
\subsubsection{More details on the design of our agent}
\keypoint{Actor}
In our experiments, the actor comprises a set of pre-defined option policies. The available options are as follows:
\begin{itemize}
\item Explore: This option is implemented using preset rules. The specific procedure is as follows: first, the agent is instructed to move to the top-left corner of the task environment, and then proceed to traverse each row alternately while scanning, until all information within the environment becomes visible. This option enables the agent to uncover unexplored areas for discovering new information.
\item Go to [an object]: With this option, the agent can navigate to an object within the environment. The object can be any interactable element, such as a key, box, or door.
\item Pickup [an object]: This option enables the agent to pick up a specified object. It is useful when the agent needs to acquire an item to progress in the task, like grabbing a key to unlock a door.
\item Toggle [an object]: Using this option, the agent can the state of a particular object. Examples include opening or closing a door, use a key to unlock a door or open a box.
\end{itemize}
These pre-defined options provide the agent with a repertoire of high-level actions to choose from during its decision-making process. By selecting the appropriate option based on its current objective and observations, the agent can efficiently navigate and interact with the environment to accomplish the given task.
For more details, refer to the supplement materials. 

\keypoint{How to train the asking policy?}
In our experiments, we train a neural network to serve as the asking policy.
Specifically, this neural network receives observations from the current and previous frames as input. Before passing these observations to the network, we compute the difference between the two frames. This encourages the asking policy to generate an ``ask" action only when there are noticeable changes in the environment compared to the previous frame.
The network architecture for the asking policy comprises three convolutional neural network (CNN) layers followed by two multilayer perceptron (MLP) layers. The output of the network consists of logits for each option, indicating the probability of selecting the ``ask" or ``not ask" action for each option. Therefore, the dimensionality of the network's output is $2\times K$, where the (2$k$-1)-th and 2$k$-th entries collectively determine the action distribution for option $k$. Here, $K$ represents the size of the option set used in our approach.
By training the asking policy network with this architecture, we enhance the agent's ability to make informed decisions regarding whether it should pose a query to the LLM planner or not, based on changes observed in the environment between consecutive frames.
\subsection{Our approach can tolerate the imperfection of the translator in the mediator module}
In the complex environment of \textit{ColoredDoorKey} in MiniGrid, the baseline interaction method \textit{Always} has been observed to fail in certain corner cases due to flaws of other components within the framework. Fig.~\ref{fig: example2} presents an example scenario in \textit{ColoredDoorKey} that showcases such a case. In the first frame, the agent is instructed to \textit{go to then pick up the key}. After taking a left turn to drop the carried purple key (frame 2), the LLM instructs the agent once again with \textit{go to then pick up the key}, where the agent should proceed to pick up the yellow key. However, the \textit{Always} baseline fails in this case because the translator does not encode information about the relative position between the agent and the target object accurately. Consequently, the translator returns the same observation [\textit{observed yellow key, observed yellow door, carrying purple key}] for both frames 1 and 2.
In contrast, our approach learns ``not to ask" for assistance in this particular case, allowing the agent to complete the action of picking up the yellow key before requesting further instructions. This highlights a significant advantage of our approach over baseline methods: it can adapt to situations where the translator's translation process loses a lot of information. The learned asking policy enables the agent to make more informed decisions based on its observations and context, leading to robust performance in scenarios where baseline methods may fail due to flaws of the translator.

\subsection{Experimental setting on Habitat}
Habitat is a simulation platform specifically designed for end-to-end development of embodied AI \cite{szot2021habitat}. It provides a framework for defining various embodied AI tasks such as navigation, object rearrangement, and question answering. Additionally, it allows for the configuration of embodied agents with specific physical forms and sensors. Agents can be trained using either imitation or reinforcement learning techniques.
In our experiments, we demonstrate that our approach can generalize effectively to visually realistic domains by conducting experiments within the Habitat environment.

In our experiments, we focus on the manipulation task known as \textit{Pick\&Place}. In this task, the robot agent's objective is to pick up an object from a desk and precisely place it into a designated target receptacle, which in this case is the kitchen sink. The task setting is the same as in the Habitat experiment in \cite{zhou2024large}.

The robot agent is equipped with a wheeled base, a 7-degree-of-freedom (DoF) arm manipulator, and a parallel-jaw gripper. Additionally, it features a camera mounted on its ``head" that provides a field of view of $90^{\circ}$ and captures visual data at a resolution of $256\times 256$ pixels. As a result, the observation space of the environment comprises a visual observation denoted as $o_v \in \mathbb{R}^{256\times 256\times 1}$ from the depth camera. It also includes a sensor observation $o_s \in \mathbb{R}^{24}$ sourced from various sensors such as joint sensors, gripping sensors, the end effector of the arm, object and target GPS sensors, among others. The action space in our setup is 11-dimensional, consisting of 3 actions controlling the robot positions, 7 actions controlling the robot arm and one action indicating termination. This action space enables the agent to execute precise movements and manipulations necessary for accomplishing the target task.

To effectively train each option, we design the rewards based on rearrangement measures. These measures take into account various factors such as the force exerted by the articulated agent, the distance between the object and the goal, and the angle between the agent and the goal. The specific details of these measures can be found in the Habitat documentations \cite{szot2021habitat}.

In the \textit{Pick\&Place} environment, as solving the task requires progressively achieving several sub-goals, we use a composite stage reward system. More specifically, picking up the object successfully is referred to as \textit{Stage1 Completion} and rewards a value of 1. Achieving navigation to the sink with the object is referred to as \textit{Stage2 Completion} and also rewards a value of 1. Finally, successfully placing the apple into the target sink is referred to as \textit{Stage3 Completion} and grants a higher reward value of 5. It is important to note that if any of the high-level options exceed their designated time limit, the task may terminate prematurely.

\subsubsection{Implementation details of our approach on Habitat}
\label{sec:imp_Habitat}
\begin{figure}
    \centering
    \includegraphics[width = 0.6\textwidth]{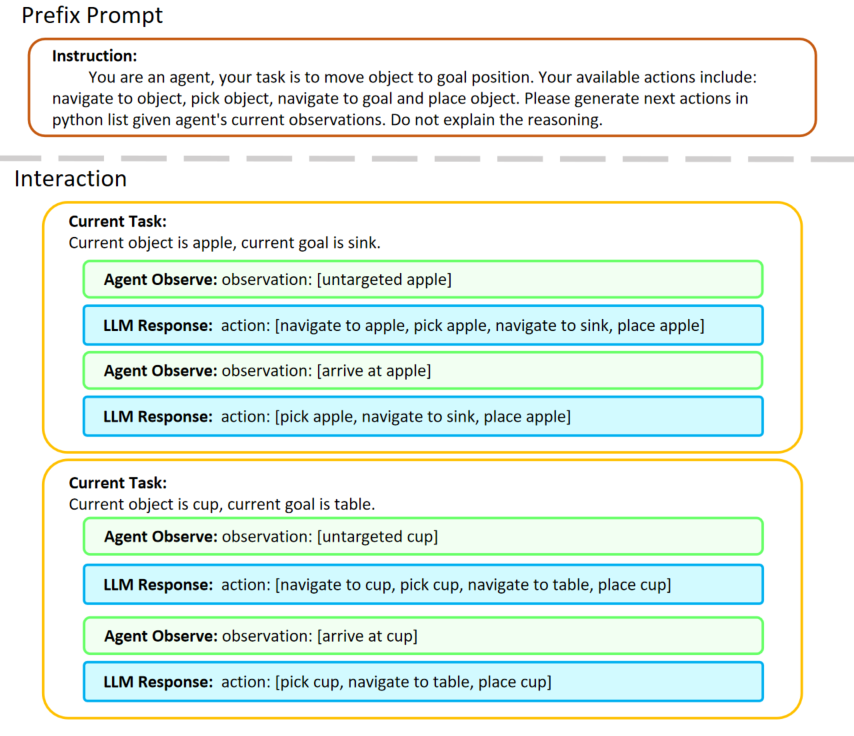}
    \caption{An example of the prompts and interactions for the habitat environment. Prefix prompt only contains a short task instruction.}
    \label{fig:prompt_habitat}
\end{figure}

\keypoint{Planner}
We employ the pre-trained Vicuna-7b model as the LLM planner in our approach. In terms of prompt design, we begin by furnishing a concise instruction that conveys information about the target task. Subsequently, we provide a description of the current observation in the form of a Python list. An example of the dialogue generated by the LLM planner can be found in Fig.~\ref{fig:prompt_habitat}.

\keypoint{Actor}
In our experiments, we use three high-level options: \{\textit{Navigate}, \textit{Pick}, \textit{Place}\}, each pre-trained with RL independently. Whenever there is a transition between these options, an automatic execution of the default action \textit{Reset Arm} occurs. To ensure effective training of these options, we use 32 distinct training environment specifications with different object locations and target locations. Additionally, the agent's initial positions are randomly generated each time the environment is reset, guaranteeing variability in training scenarios. For each option, we employ a ResNet18 backbone combined with a 2-layer LSTM architecture to train the corresponding models. During testing, the success rates of \textit{Navigate}, \textit{Pick}, and \textit{Place} are 84\%, 92\%, and 91\% respectively. These pre-trained models remain fixed throughout the task, ensuring consistency and stability during execution.

\keypoint{Training of the asking policy}
Similar to our Minigrid experiment, we stack five consecutive frames of observations as inputs to the asking policy. This enables the network to capture temporal information and make informed decisions based on past observations. The network architecture for the asking policy consists of three CNN layers for embedding visual observations, one MLP layer for embedding sensor observations, and two additional MLP layers to output the logits for the binary question of whether to \textit{ask} or \textit{not ask}.
\end{document}